\documentclass[runningheads]{llncs}
\usepackage{graphicx}
\usepackage{amsmath,amssymb} 
\usepackage{color}
\usepackage{epstopdf}
\usepackage[width=122mm,left=12mm,paperwidth=146mm,height=193mm,top=12mm,paperheight=217mm]{geometry}
\usepackage{epsfig,amsmath,amssymb,epsf,algorithm,cite,algorithmic,adjustbox}
\usepackage{subcaption}
\DeclareMathOperator*{\argminA}{argmin}   
\DeclareMathOperator*{\argmaxA}{argmax}   

\graphicspath{{./figure/}}

 \def\btheta{{\pmb{\theta}}}

\def\b0{{\pmb{0}}}

   \def\bx{{\boldsymbol{x}}}
\def\be{{\boldsymbol{e}}}   
   \def\bz{{\boldsymbol{z}}}

\begin{document}
\pagestyle{headings}
\mainmatter

\title{Fictitious GAN: Training GANs with Historical Models} 

\titlerunning{Fictitious GAN}

\authorrunning{Hao Ge, Yin Xia, Xu Chen, Randall Berry, Ying Wu}

\author{ Hao Ge, Yin Xia, Xu Chen, Randall Berry, Ying Wu \thanks{First three authors have equal contributions.}}
\institute{Northwestern University, Evanston, IL, USA \\
	\{haoge2013,yinxia2012,chenx\}@u.northwestern.edu \\
	\{rberry,yingwu\}@northwestern.edu}

\maketitle

\begin{abstract}
	Generative adversarial networks (GANs) are powerful tools for learning generative models.  In practice, the training may suffer from lack of convergence. GANs are commonly viewed as a two-player zero-sum game between two neural networks. Here, we leverage this game theoretic view to study the convergence behavior of the training process. Inspired by the fictitious play learning process, a novel training method, referred to as Fictitious GAN, is introduced. Fictitious GAN trains the deep neural networks using a mixture of historical models. Specifically, the discriminator (resp. generator) is updated according to the best-response to the \textit{mixture} outputs from a sequence of previously trained generators (resp. discriminators). It is shown that Fictitious GAN can effectively resolve some convergence issues that cannot be resolved by the standard training approach. It is proved that asymptotically the average of the generator outputs has the same distribution as the data samples. 
\end{abstract}

\section{Introduction}

\subsection{Generative Adversarial Networks}

Generative adversarial networks (GANs) are a powerful framework for learning generative models. They have witnessed successful applications in a wide range of fields, including image synthesis~\cite{reed2016generative,shrivastava2016learning}, image super-resolution~\cite{johnson2016perceptual,ledig2016photo}, and anomaly detection~\cite{zhai2016deep}. A GAN maintains two deep neural networks: the discriminator and the generator. The generator aims to produce samples that resemble the data distribution, while the discriminator aims to distinguish the generated samples and the data samples.

Mathematically, the standard GAN training aims to solve the following optimization problem:
\begin{align}\label{eq:gan}
\min_{G} \max_{D} V(G,D) = \mathsf{E}_{\bx \sim p_d(\bx)} \{\log D(\bx) \} + \mathsf{E}_{\bz \sim p_z(\bz)} \{\log (1- D(G(\bz)) ) \}.
\end{align}
The global optimum point is reached when the generated distribution $p_g$, which is the distribution of $G(\bz)$ given $\bz \sim p_z(\bz)$, is equal to the data distribution. 
The optimal point is reached based on the assumption that the discriminator and generator are \textit{jointly} optimized. 
Practical training of GANs, however, may not satisfy this assumption. 
In some training process, instead of ideal joint optimization, the discriminator and generator seek for best response by turns, namely the discriminator (resp. generator) is alternately updated with the generator (resp. discriminator) fixed.


Another conventional training methods are based on a gradient descent form of GAN optimization. In particular, they simultaneously take small gradient steps in both generator and discriminator parameters in each training iteration~\cite{goodfellow2014generative}. There have been some studies on the convergence behaviors of gradient-based training. The local convergence behavior has been studied in~\cite{nagarajan2017gradient,heusel2017gans}. The gradient-based optimization is proved to converge assuming that the discriminator and the generator is convex over the network parameters~\cite{nowozin2016f}. The inherent connection between gradient-based training and primal-dual subgradient methods for solving convex optimizations is built in~\cite{chen2018training}.

Despite the promising practical applications, a lot of works still witness the lack of convergence behaviors in training GANs. Two common failure modes are oscillation and mode collapse, where the generator only produces a small family of samples \cite{goodfellow2014generative,li2017towards,che2016mode}. One important observation in~\cite{metz2016unrolled} is that such non convergence behaviors stem from the fact that each generator update step is a partial collapse towards a delta function, which is the best response to the objective function. This motivates the study of this paper on the dynamics of best-response training and the proposal of a novel training method to address these convergence issues. 




\subsection{Contributions}

In this paper, we view GANs as a two-player zero-sum game and the training process as a repeated game. For the optimal solution to Eq. \eqref{eq:gan}, the corresponding generated distribution and discriminator $(p^*_g, D^*)$ is shown to be the unique Nash equilibrium in the game. 
Inspired by the well-established fictitious play mechanism in game theory, we propose a novel training algorithm to resolve the convergence issue and find this Nash equilibrium.

The proposed training algorithm is referred to as Fictitious GAN, where the discriminator (resp. generator) is updated based on the the mixed outputs from the sequence of historical trained generators (resp. discriminators). 
The previously trained models actually carry important information and can be utilized for the updates of the new model. 
We prove that Fictitious GAN achieves the optimal solution to Eq. \eqref{eq:gan}. In particular, the discriminator outputs converge to the optimum discriminator function and the mixed output from the sequence of trained generators converges to the data distribution. 

Moreover, Fictitious GAN can be regarded as a meta-algorithm that can be applied on top of existing GAN variants. Both synthetic data and real-world image datasets are used to demonstrate the improved performance due to the fictitious training mechanism. 

\section{Related Works}

The idea of  training using multiple GAN models have been considered in other works. In \cite{arora2017generalization,hoang2017multi}, the mixed outputs of multiple generators is used to approximate the data distribution. The multiple generators with a modified loss function have been used to alleviate the mode collapse problem~\cite{ghosh2017multi}.
In \cite{metz2016unrolled}, the generator is updated based on a sequence of unrolled discriminators. In \cite{nguyen2017dual}, dual discriminators are used to combine the Kullback-Leibler (KL) divergence and reverse KL divergences into a unified objective function. Using an ensemble of discriminators or GAN models has shown promising performance~\cite{durugkar2016generative,tolstikhin2017adagan}. One distinguishing difference between the above-mentioned methods and our proposed method is that in our method only a \textit{single} deep neural network is trained at each training iteration, while multiple generators (resp. discriminators) only provide inputs to a single discriminator (resp. generators) at each training stage.  Moreover, the outputs from multiple networks is simply uniformly averaged and serves as input to the target training network, while other works need to train the optimal weights to average the network models. The proposed method thus has a much lower computational complexity. 

The use of historical models have been proposed as a heuristic method to increase the diversity of generated samples~\cite{salimans2016improved}, while the theoretical convergence guarantee is lacking. Game theoretic approaches have been utilized to achieve a resource-bounded Nash equilibrium in GANs~\cite{oliehoek2018beyond}.
Another closely related work to this paper is the recent work~\cite{grnarova2017online} that applies the Follow-the-Regularized-Leader (FTRL) algorithm to train GANs. In their work, the historical models are also utilized for online learning. There are at least two distinct features in our work. First, we borrow the idea of fictitious play from game theory to prove convergence to the Nash equilibrium for any GAN architectures assuming that networks have enough capacity, while \cite{grnarova2017online} only proves convergence for semi-shallow architectures. Secondly, we prove that a \textit{single} discriminator, instead of a mixture of multiple discriminators, asymptotically converges to the optimal discriminator. This provides important design guidelines for the training, where asymptotically a single discriminator needs to be maintained. \footnote{Due to space constraints, all the proofs in the paper are omitted and can be found in the Supplementary materials.}

\section{Toy Examples}
\label{sec:toy}

In this section, we use two toy examples to show that both the best-response approach and the gradient-based training approach may oscillate for simple minimax optimization problems. 

Take the GAN framework for instance, for the best-response training approach, the discriminator and the generator are updated to the optimum point at each iteration. Mathematically, the discriminator and the generator is alternately updated according to the following rules:
\begin{align}
\label{maxD}& \max_{D} \mathsf{E}_{\bx \sim p_d(\bx)} \{\log D(\bx) \} + \mathsf{E}_{\bz \sim p_z(\bz)} \{\log (1- D(G(\bz)) \} \\
& \min_{G} \mathsf{E}_{\bz \sim p_z(\bz)} \{\log (1- D( G(\bz) )) \} \label{gloss}
\end{align}

%
%

\begin{example}~\label{example:oscillate}
	Let the data follow the Bernoulli distribution $p_d \sim $ Bernoulli $(a)$, where $0 <a <1$. Suppose the initial generated distribution $p_g \sim$ Bernoulli $(b)$, where $b \neq a$. We show that in the best-response training process, the generated distribution oscillates between $p_g \sim$ Bernoulli $(1)$ and $p_g \sim$ Bernoulli $(0)$.
\end{example}

We show the oscillation phenomenon in training using best-response training approach. To minimize \eqref{gloss}, it is equivalent to find $p_g$ such that $ \mathsf{E}_{\bx \sim p_g(\bx)} \{\log (1- D(\bx )) \}$ is minimized. At each iteration, the output distribution of the updated generator would concentrate all the probability mass at $x=0$ if $D(0) > D(1)$, or at $x=1$ if $D(0) < D(1)$. Suppose $p_g(x) = 1\{x=0\}$, where $1\{\cdot\}$ is the indicator function, then by solving (\ref{maxD}), the discriminator at the next iteration is updated as
\begin{align}
D(x) = \frac{p_d (x)}{p_d(x) + p_g (x)},
\end{align}
which yields $D(1) =1$ and $D(0) < D(1)$. Therefore, the generated distribution at the next iteration becomes $p_g(x) = 1\{x = 1\}$. The oscillation between $p_g \sim$ Bernoulli $(1)$ and $p_g \sim$ Bernoulli $(0)$ continues by induction. A similar phenomenon can be observed for Wasserstein GAN.

The first toy example implies that the oscillation behavior is a fundamental problem to the iterative best-response training. In practical training of GANs, instead of finding the best response, the discriminator and generator are updated based on gradient descent towards the best-response of the objective function. However, the next example adapted from~\cite{goodfellow2016nips} demonstrates the failure of convergence in a simple minimax problem using a gradient-based method.

\begin{example}\label{example:vxy}
	Consider the following minimax problem:
	\begin{align}
	\min_{-10 \leq y \leq 10} \max_{-10 \leq x \leq 10} xy.
	\end{align}
\end{example}

Consider the gradient based training approach with step size $\triangle$. The update rule of $x$ and $y$ is:
\begin{equation}\label{eq:trans}
\begin{bmatrix}
x_{n+1}    \\
y_{n+1}  
\end{bmatrix}
=
\begin{bmatrix}
1 & \triangle  \\
-\triangle & 1
\end{bmatrix}
\begin{bmatrix}
x_{n}    \\
y_{n}  
\end{bmatrix}.
\end{equation}
By using the knowledge of eigenvalues and eigenvectors, we can obtain
\begin{equation}\label{eq:dynamic_example2}
\begin{bmatrix}
x_{n}    \\
y_{n}  
\end{bmatrix}
=
\begin{bmatrix}
-c_1^nc_2\sin(n\theta+\beta)    \\
c_1^nc_2\cos(n\theta+\beta)   
\end{bmatrix},
\end{equation}
where $c_1 = \sqrt{1+\triangle^2} > 1$ and $c_2,\theta,\beta$ are constants depending on the initial $(x_0,y_0)$. As $n\rightarrow \infty$, since $c_1>1$, the process will not converge.
\begin{figure}[t]
	\centering
	\begin{subfigure}[b]{0.3\textwidth}
		\centering
		\includegraphics[width=1.0\textwidth]{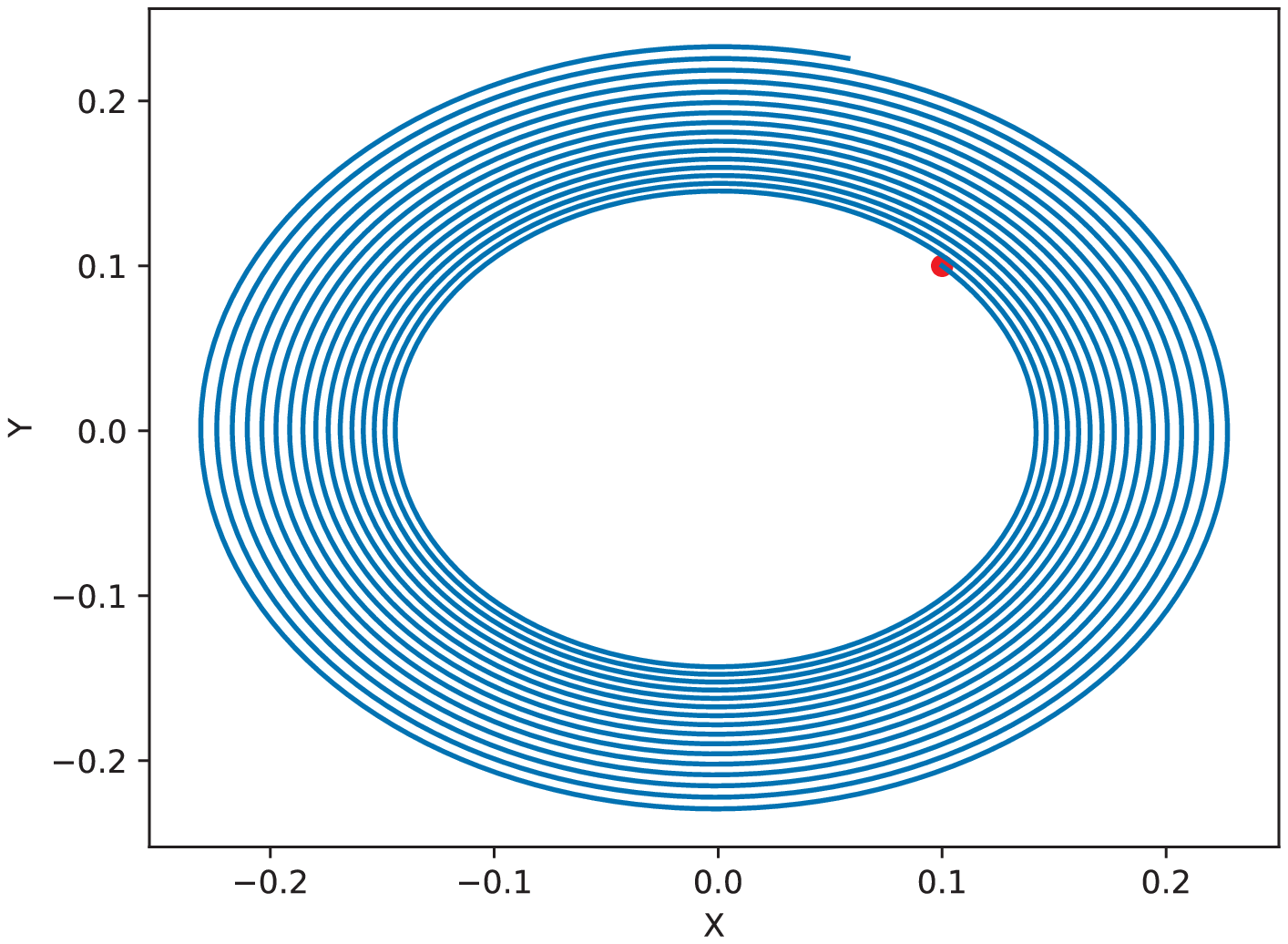}
		\caption{}
		\label{fig:curve}
	\end{subfigure}
	\begin{subfigure}[b]{0.3\textwidth}
		\centering
		\includegraphics[width=1.0\textwidth]{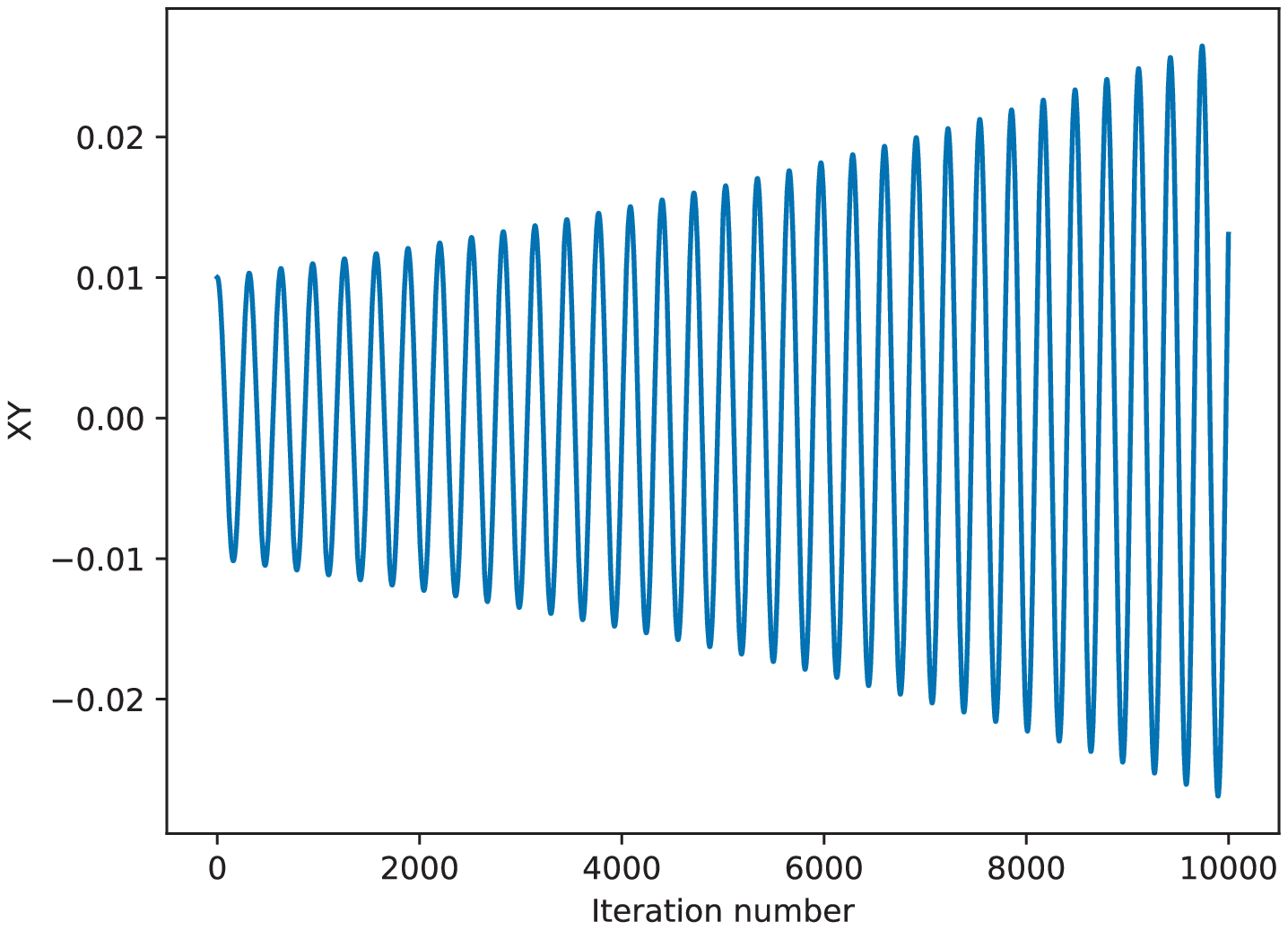}
		\caption{}
		\label{fig:payoff}
	\end{subfigure}
	\caption{Performance of gradient method with fixed step size for Example~\ref{example:vxy}. (a) illustrates the choices of x and y as iteration processes, the red point $(0.1,0.1)$ is the initial value. (b) illustrates the value of $xy$ as a function of iteration numbers.  }
	\label{fig:gradient}
\end{figure}

Figure \ref{fig:gradient} shows the performance of gradient based approach, the initial value $(x_0,y_0) = (0.1,0.1)$ and step size is 0.01. It can be seen that both players' actions do not converge. This toy example shows that even the gradient based approach with arbitrarily small step size may not converge.


We will revisit the convergence behavior in the context of game theory. A well-established learning mechanism in game theory naturally leads to a training algorithm that resolves the non-convergence issues of these two toy examples.
\section{Nash Equilibrium in Zero-Sum Games}

In this section, we introduce the two-player zero-sum game and describe the learning mechanism of fictitious play, which provably achieves a Nash equilibrium of the game. We will show that the minimax optimization of GAN can be formulated as a two-player zero-sum game, where the optimal solution corresponds to the unique Nash equilibrium in the game. In the next section we will propose a training algorithm which simulates the fictitious play mechanism and provably achieves the optimal solution.

\subsection{Zero-Sum Games}

We start with some definitions in game theory. A game consists of a set of $n$ players, who are rational and take actions to maximize their own utilities. Each player $i$ chooses a pure strategy $s_i$ from the strategy space $S_i = \{s_{i,0},\cdots,s_{i,m-1}\}$. Here player $i$ has $m$ strategies in her strategy space. 
A utility function $u_i(s_i,s_{-i})$, which is defined over all players' strategies, indicates the outcome for player $i$, where the subscript $-i$ stands for all players excluding player $i$. There are two kinds of strategies, pure and mixed strategy. A pure strategy provides a specific action that a player will follow for any possible situation in a game, while a mixed strategy $\mu_i=(p_i(s_{i,0}),\cdots,p_i(s_{i,m-1}))$ for player $i$ is a probability distribution over the $m$ pure strategies in her strategy space with $\sum_j p_i(s_{i,j}) =1$. The set of possible mixed strategies available to player $i$ is denoted by $\Delta S_i$. The expected utility of mixed strategy $(\mu_i,\mu_{-i})$ for player $i$ is 
\begin{equation}
\mathsf{E} \left\{ u_i(\mu_i,\mu_{-i}) \right\}= \sum_{s_i \in S_i}\sum_{s_{-i} \in S_{-i}} u_i(s_i,s_{-i})p_i(s_i)p_{-i}(s_{-i}).
\end{equation}

For ease of notation, we write $u_i(\mu_i,\mu_{-i})$ as $\mathsf{E} \left\{ u_i(\mu_i,\mu_{-i}) \right\}$ in the following.
Note that a pure strategy can be expressed as a mixed strategy that places probability 1 on a single pure strategy and probability 0 on the others. A game is referred to as a finite game or a continuous game, if the strategy space is finite or nonempty and compact, respectively. In a continuous game, the mixed strategy indicates a probability density function (pdf) over the strategy space.

\begin{definition}
	For player i, a strategy $\mu_i^*$ is called a best response to others' strategy $\mu_{-i}$ if $u_i(\mu_i^*,\mu_{-i})\geq u_i(\mu_i,\mu_{-i})$ for any $\mu_i \in \Delta S_i.$
\end{definition}

\begin{definition}
	A set of mixed strategies $\mu^{\ast} = (\mu_1^{\ast}, \mu_2^{\ast}, \cdots, \mu_n^{\ast})$ is a Nash equilibrium if, for every player $i$, $\mu_i^*$ is a best response to the strategies $\mu_{-i}^*$ played by the other players in this game.
\end{definition}




\begin{definition}
	A zero-sum game is one in which each player's gain or loss is exactly balanced by the others' loss or gain and the sum of the players' payoff is always zero.
\end{definition}
Now we focus on a continuous two-player zero-sum game. In such a game, given the strategy pair $(\mu_1, \mu_2)$, player 1 has a utility of $u(\mu_1, \mu_2)$, while player 2 has a utility of $-u(\mu_1, \mu_2)$. In the framework of GAN, the training objective \eqref{eq:gan} can be regarded as a two-player zero-sum game, where the generator and discriminator are two players with utility functions $-V(G,D)$ and $V(G,D)$, respectively. Both of them aim to maximize their utility and the sum of their utilities is zero.  

Knowing the opponent is always seeking to maximize its utility, Player 1 and 2 choose strategies according to
\begin{align}
\mu_1^* &= \argmaxA_{\mu_1 \in \Delta S_1} \min_{\mu_2 \in \Delta S_2} u(\mu_1,\mu_2) \\
\mu_2^* &= \argminA_{ \mu_2 \in \Delta S_2 }\max_{\mu_1 \in \Delta S_1} u(\mu_1,\mu_2).
\end{align} 

Define $\underline{v}=\max\limits_{\mu_1\in \Delta S_1}\min\limits_{\mu_{2}\in \Delta S_2} u(\mu_1,\mu_{2})$ and $\bar{v}=\min\limits_{\mu_{2}\in \Delta S_{2}}\max\limits_{\mu_{1}\in \Delta S_{1}} u (\mu_1,\mu_{2})$ as the lower value and upper value of the game, respectively. Generally, $\underline{v} \leq \bar{v}$. Sion~\cite{sion1958general} showed that these two values coincide under some regularity conditions:

\begin{theorem}[Sion's Minimax Theorem~\cite{sion1958general}]\label{theorem:sion}
	Let $X$ and $Y$ be convex, compact spaces, and $f$: $X \times Y \rightarrow \mathbb R$. If for any $x \in X$, $f(x,\cdot)$ is upper semi-continuous and quasi-concave on $Y$ and for any $y\in Y$, $f(\cdot,y)$ is lower semi-continuous and quasi-convex on $X$, then $\inf_{x\in X}\sup_{y\in Y}f(x,y) = \sup_{y\in Y}\inf_{x\in X}f(x,y)$.
\end{theorem}

Hence, in a zero-sum game, if the utility function $u(\mu_1,\mu_2)$ satisfies the conditions in Theorem~\ref{theorem:sion}, then $\underline{v} = \bar{v}$. We refer to $v = \underline{v} = \bar{v}$ as the\textit{ value of the game}. We further show that a Nash equilibrium of the zero-sum game achieves the value of the game. 
\begin{corollary}\label{cor:ne_minimax}
	In a two-player zero-sum game with the utility function satisfying the conditions in Theorem~\ref{theorem:sion}, if a strategy $(\mu_1^*,\mu_2^*)$ is a Nash equilibrium, then $u(\mu_1^*,\mu_2^*) = v$.
\end{corollary}


Corollary \ref{cor:ne_minimax} implies that if we have an algorithm that achieves a Nash equilibrium of a zero-sum game, we may utilize this algorithm to optimally train a GAN. We next describe a learning mechanism to achieve a Nash equilibrium.

\subsection{Fictitious Play}
Suppose the zero-sum game is played repeatedly between two rational players, then each player may try to infer her opponent's strategy. Let $s_i^{n} \in S_i$ denote the action taken by player $i$ at time $n$. At time $n$, given the previous actions $\{s_{2}^0,s_{2}^1,\cdots,s_{2}^{n-1}\}$ chosen by player $2$, one good hypothesis is that player $2$ is using stationary mixed strategies and chooses strategy $s_{2}^t$, $0\leq t\leq n-1$, with probability $\frac{1}{n}$. Here we use the empirical frequency to approximate the probability in mixed strategies. Under this hypothesis, the best response for player $1$ at time $n$ is to choose the strategy $\mu_1^*$ satisfying:
\begin{equation}
\mu_1^* =\argmaxA_{\mu_1\in \Delta S_1} u (\mu_1,\mu_2^n),
\end{equation}
where $\mu_2^n$ is the empirical distribution of player 2's historical actions. Similarly, player 2 can choose the best response assuming player 1 is choosing its strategy according to the empirical distribution of the historical actions.

Notice that the expected utility is a linear combination of utilities under different pure strategies, hence for any hypothesis $\mu_{-i}^n$, player $i$ can find a pure strategy $s_i^{n}$ as a best response. Therefore, we further assume each player plays the best pure response at each round. In game theory this learning rule is called \textit{fictitious play}, proposed by Brown \cite{brown1951iterative}.
%
%


Danskin \cite{danskin1954fictitious} showed that for any continuous zero-sum games with any initial strategy profile, fictitious play will converge. This important result is summarized in the following theorem.
\begin{theorem}\label{theorem:fict_play_game}
	Let $u(s_1,s_2)$ be a continuous function defined on the direct product of two compact sets $S_1$ and $S_2$. The pure strategy sequences $\{s_1^n\}$ and $\{s_2^n\}$ are defined as follows: $s_1^0$ and $s_2^0$ are arbitrary, and 
	\begin{equation}
	s_1^n \in \argmaxA_{s_1\in S_1} \frac{1}{n} \sum_{k=0}^{n-1}u (s_1,s_2^k), \ \ \ \ s_2^n \in \argminA_{s_2\in S_2}\frac{1}{n} \sum_{k=0}^{n-1} u (s_1^k,s_2),
	\end{equation}
	then 
	\begin{equation}
	\lim_{n\rightarrow \infty}\frac{1}{n}\sum_{k=0}^{n-1} u (s_1^n,s_2^k)=\lim_{n\rightarrow \infty}\frac{1}{n}\sum_{k=0}^{n-1}u (s_1^k,s^n_2)=v,
	\end{equation}
	where $v$ is the value of the game. 	 
\end{theorem}

\subsection{Effectiveness of Fictitious Play}

In this section, we show that fictitious play enables the convergence of learning to the optimal solution for the two counter-examples in Section~\ref{sec:toy}.

\begin{figure}[t]
	\centering
	\begin{subfigure}[b]{0.32\textwidth}
		\centering
		\includegraphics[width=1\textwidth]{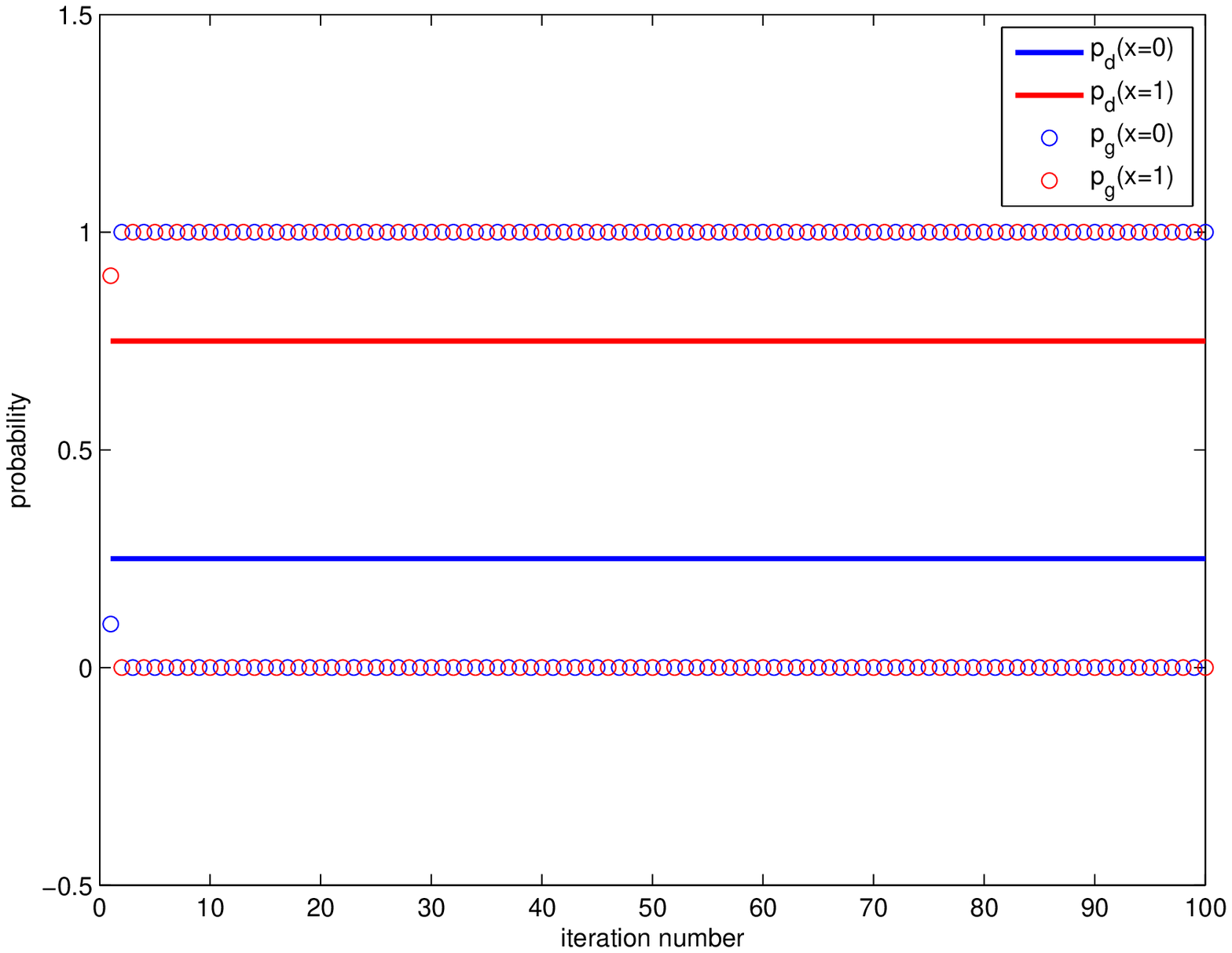}
		\caption{}
		\label{fig:fict_example1_a}
	\end{subfigure}
	\begin{subfigure}[b]{0.32\textwidth}
		\centering
		\includegraphics[width=1\textwidth]{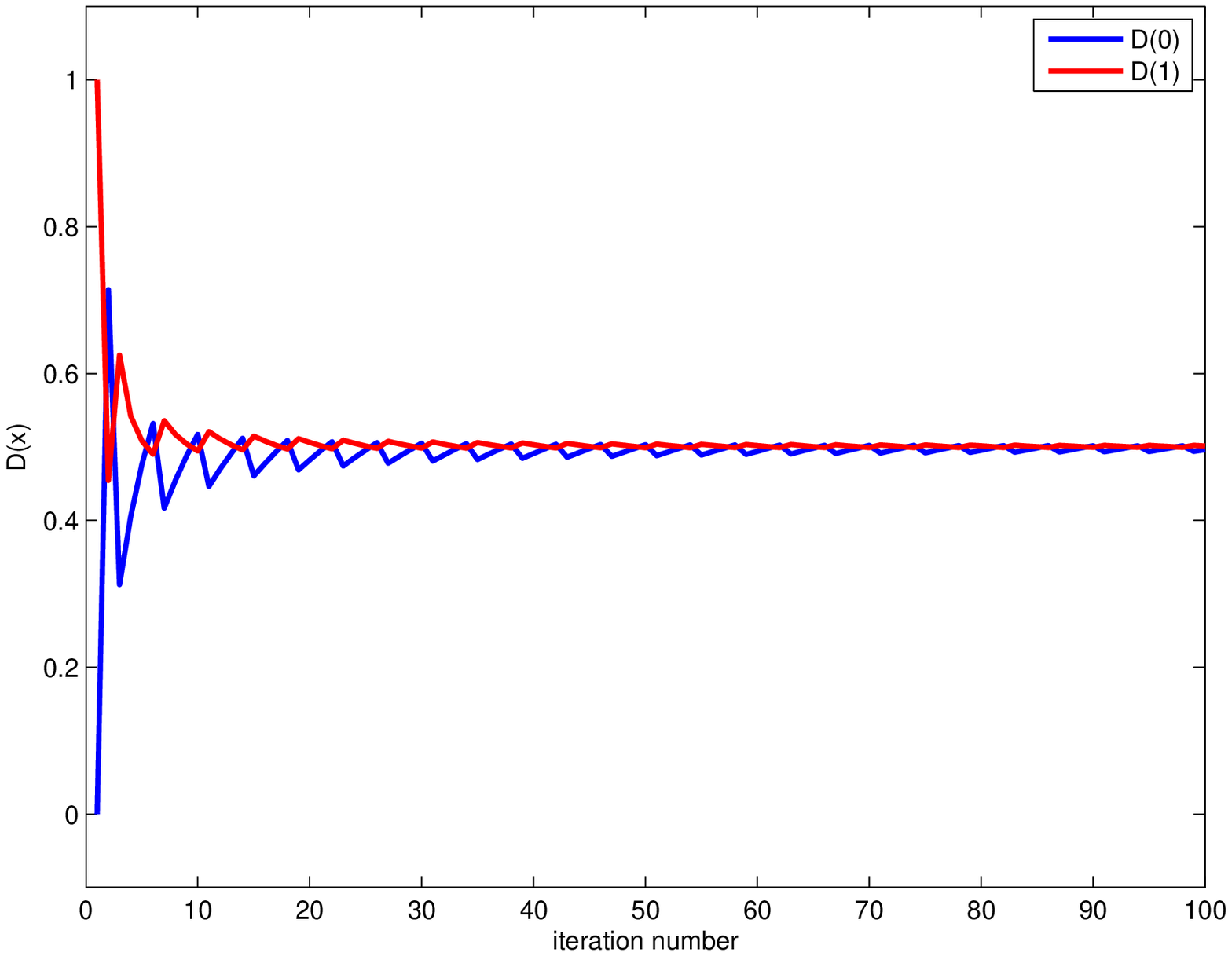}
		\caption{}
	\end{subfigure}
	\begin{subfigure}[b]{0.32\textwidth}
		\centering
		\includegraphics[width=1\textwidth]{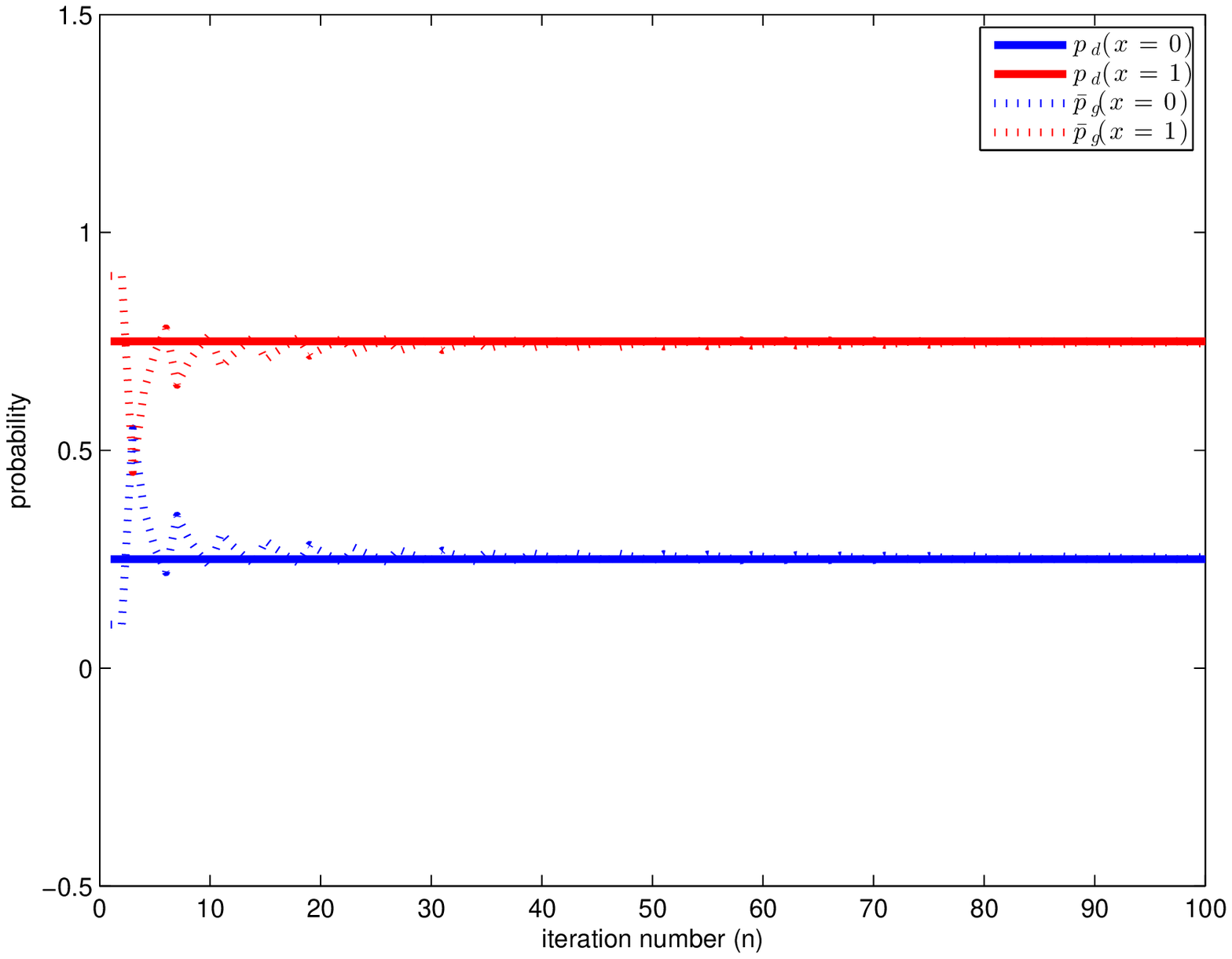}
		\caption{}
	\end{subfigure}
	\caption{Performance of best-response training for Example~\ref{example:oscillate}. (a) is Bernoulli distribution of $p_g$ assuming best-response updates. (b) illustrates $D(x)$ in Fictitious GAN assuming best response at each training iteration. (c) illustrates the average of $p_g(x)$ in Fictitious GAN assuming best response at each training iteration.}
	\label{fig:fict_example1}
\end{figure}

\textbf{Example~\ref{example:oscillate}:} 
Fig.~\ref{fig:fict_example1} shows the performance of the best-response approach, where the data follows a Bernoulli distribution $p_d \sim $ Bernoulli $(0.25)$, the initialization is $D(x) = x$ for $x \in [0,1]$ and the initial generated distribution $p_g \sim $ Bernoulli $(0.1)$. It can be seen that the generated distribution based on best responses oscillates between $p_g(x=0) = 1$ and $p_g(x=1) = 1$.

Assuming best response at each iteration $n$, under fictitious play, the discriminator is updated according to $D_{n} = \arg\max_{D} \frac{1}{n} \sum_{w=0}^{n-1} V(p_{g,w}, D)$ and the generated distribution is updated according to $p_{g,n} = \arg\max_{p_g}\frac{1}{n} \sum_{w=0}^{n-1} V(p_g, D_w)$. Fig~\ref{fig:fict_example1} shows the change of $D_n$ and the empirical mean of the generated distributions $\bar{p}_{g,n} = \frac{1}{n} \sum_{w=0}^{n-1} p_{g,w}$ as training proceeds. Although the best-response generated distribution at each iteration oscillates as in Fig.~\ref{fig:fict_example1_a}, the learning mechanism of fictitious play makes the empirical mean $\bar{p}_{g,n}$ converge to the data distribution. 

\begin{figure}[htb]
	\centering
	\begin{subfigure}[b]{0.32\textwidth}
		\centering
		\includegraphics[width=1.0\textwidth]{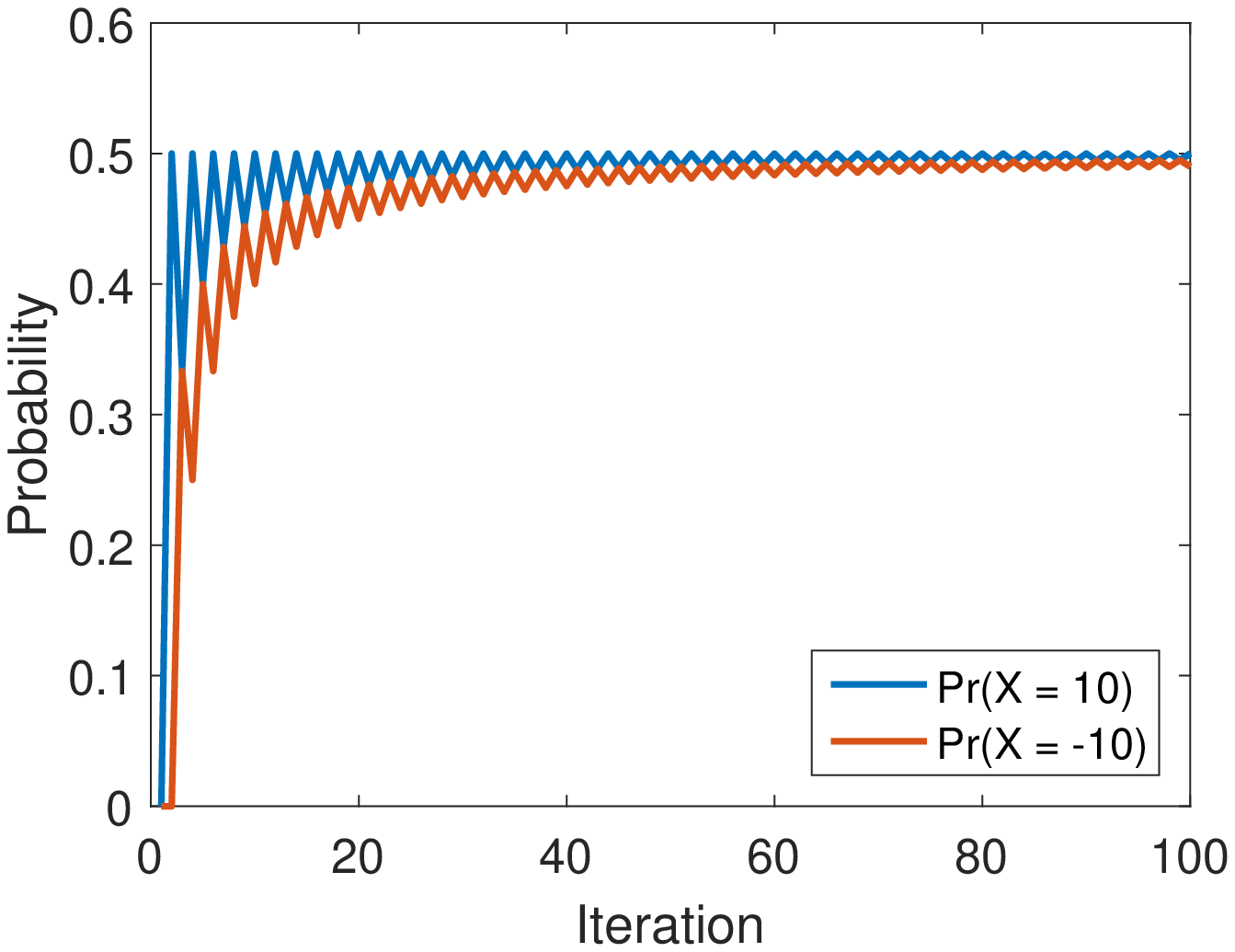}
		\caption{}
		\label{fig:it10}
	\end{subfigure}
	\begin{subfigure}[b]{0.32\textwidth}
		\centering
		\includegraphics[width=1.0\textwidth]{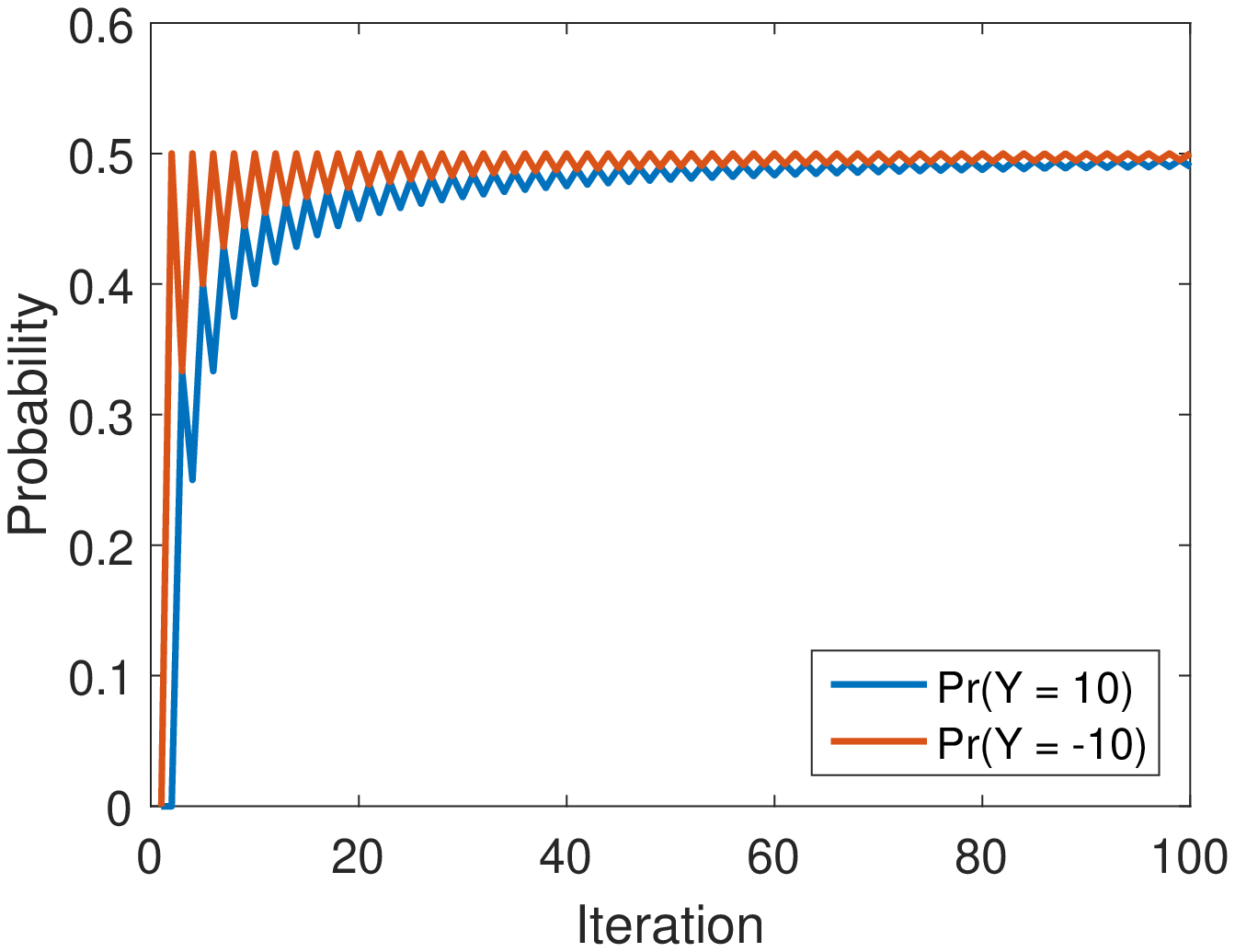}
		\caption{}
		\label{fig:it50}
	\end{subfigure}
	\begin{subfigure}[b]{0.32\textwidth}
		\centering
		\includegraphics[width=1.0\textwidth]{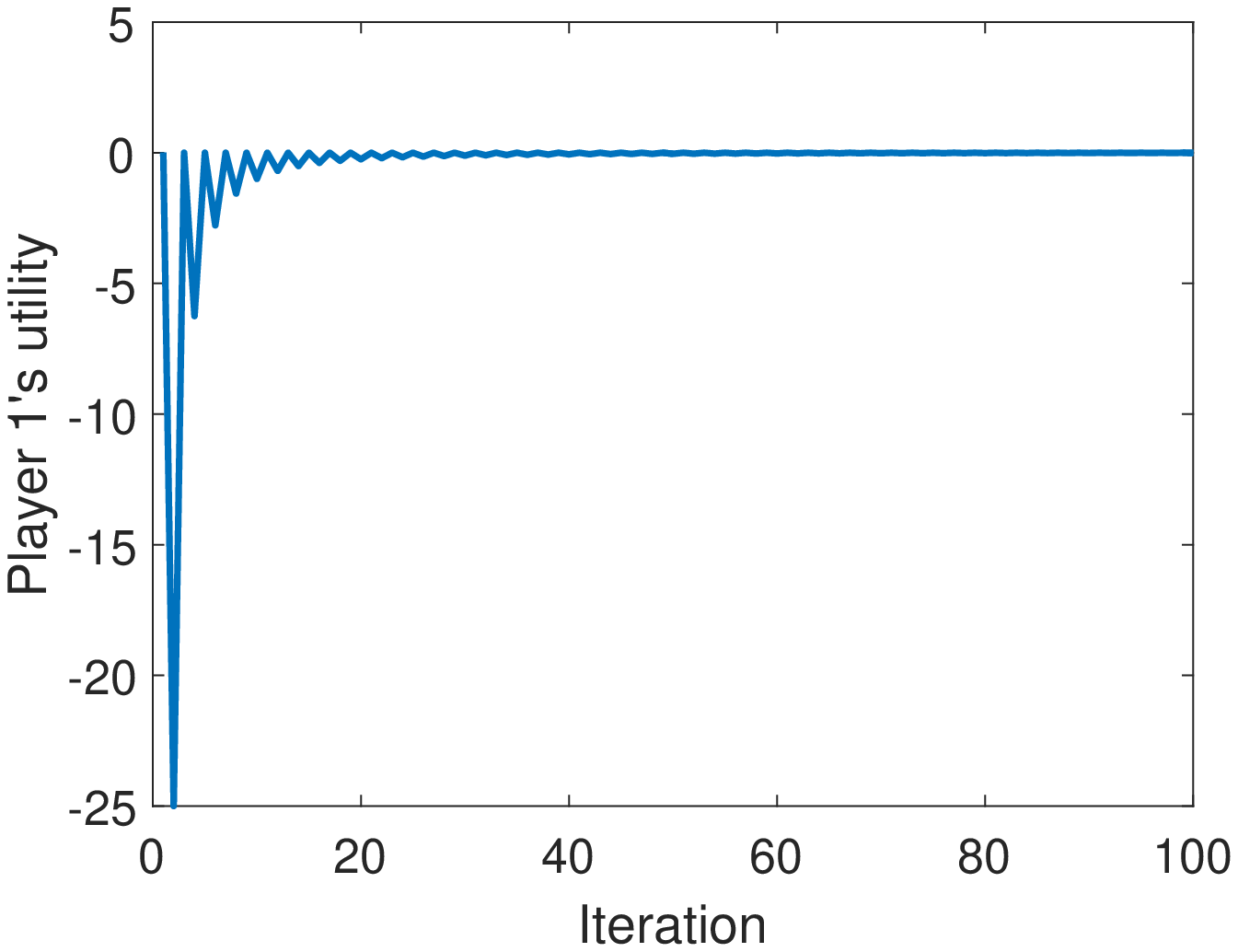}
		\caption{}
		\label{fig:mean}
	\end{subfigure}\\
	\caption{(a) and (b) illustrate the empirical distribution of $x$ and $y$ at 10 and -10, respectively. (c) illustrates the expected utility for player 1 under fictitious play. }
	\label{fig:toy_2fic}
\end{figure}

\textbf{Example~\ref{example:vxy}:} At each iteration $n$, player 1 chooses 
$x = \arg \max_x \frac{1}{n}\sum_{i=0}^{n-1} x y_i$, which is equal to $ 10* \text{sign}(\sum_{i=0}^{n-1}y_i)$.
Similarly, player 2 chooses $y$ according to $y = -10* \text{sign}(\sum_{i=0}^{n-1}x_i)$. Hence regardless of what the initial condition is, both players will only choose 10 or -10 at each iteration. Consequently, as iteration goes to infinity, the empirical mixed strategy only proposes density on 10 and -10. It is proved in the Supplementary material that the mixed strategy $(\sigma_1^*,\sigma_2^*)$ that both players choose 10 and -10 with probability $\frac{1}{2}$ is a Nash equilibrium for this game. Fig \ref{fig:toy_2fic} shows that under fictitious play, both players' empirical mixed strategy converges to the Nash equilibrium and the expected utility for each player converges to 0.

One important observation is fictitious play can provide the Nash equilibrium if the equilibrium is unique in the game. However, if there exist multiple Nash equilibriums, different initialization may yield different solutions. In the above example, it is easy to check $(0,0)$ is also a Nash equilibrium, which means both players always choose 0, but fictitious play can lead to this solution only when the initialization is $(0,0)$. The good thing we show in the next section is, due to the special structure of GAN (the utility function is linear over generated distribution), fictitious play can help us find the desired Nash equilibrium.

\section{Fictitious GAN}

\subsection{Algorithm Description}

As discussed in the last section, the competition between the generator and discriminator in GAN can be modeled as a two-player zero-sum game. The following theorem proved in the supplementary material shows that the optimal solution of \eqref{eq:gan} is actually a unique Nash equilibrium in the game. 
\begin{theorem}\label{theorem:gan_ne}
	Consider \eqref{eq:gan} as a two-player zero-sum game. The optimal solution of \eqref{eq:gan} with $p^*_g=p_{d}$ and $D^*({\bx})=1/2$ is a unique Nash equilibrium in this game. The value of the game is $-\log 4$.
\end{theorem}

By relating GAN with the two-player zero-sum game, we can design a training algorithm to simulate the fictitious play  such that the training outcome converges to the Nash equilibrium 

Fictitious GAN, as described in Algorithm \ref{alg1}, adapts the fictitious play learning mechanism to train GANs. 
We use two queues $\mathcal{D}$ and $\mathcal{G}$ to store the historically trained models of the discriminator and the generator, respectively. 
At each iteration, the discriminator (resp. generator) is updated according to the best response to $V(G,D)$ assuming that the generator (resp. discriminator) chooses a historical strategy uniformly at random. Mathematically, the discriminator and generator are updated according to \eqref{eq:mix_g} and \eqref{eq:mix_d}, where the outputs due to the generator and the discriminator is mixed uniformly at random from the previously trained models. Note the the back-propagation is still performed on a single neural network at each training step.
Different from standard training approaches, we perform $k_0$ gradient descent updates when training the discriminator and the generator in order to achieve the best response. In practical learning, queues  $\mathcal{D}$ and $\mathcal{G}$ are maintained with a fixed size. The oldest model is discarded if the queue is full when we update the discriminator or the generator.
\begin{algorithm}
	\caption{Fictitious GAN training algorithm. }\label{alg1}
	\begin{algorithmic}
		\STATE \textbf{Initialization:} Set $\mathcal{D}$ and $\mathcal{G}$ as the queues to store the historical models of the discriminators and the generators, respectively. 
		\WHILE {the stopping criterion is not met}
		\FOR {$k=1,\cdots, k_0$}
		\STATE Sample data via minibatch $\bx_1, \cdots, \bx_m$. 
		\STATE Sample noise via minibatch $\bz_1, \cdots, \bz_m$.
		\STATE Update the discriminator via gradient ascent:
		\begin{align}\label{eq:mix_g}
		\nabla_{\btheta_d} \frac{1}{m} \sum\limits_{i=1}^m \left[\ \log (D(\bx_i)) + \frac{1}{|\mathcal{G}| } \sum\limits_{G_w \in \mathcal{G}}  \log (1 - D(G_w (\bz_i )  )) \right].
		\end{align}
		\ENDFOR
		\FOR {$k=1,\cdots, k_0$}
		\STATE Sample noise via minibatch $\bz_1, \cdots, \bz_m$.
		\STATE Update the generator via gradient descent:
		\begin{align}\label{eq:mix_d}
		\nabla_{\btheta_g} \left[ \frac{1}{m |\mathcal{G}| }\sum\limits_{i=1}^m \sum\limits_{D_w \in \mathcal{D}}  \log (1 - D_w (G(\bz_i) ))  \right].
		\end{align}
		\ENDFOR
		\STATE {Insert the updated discriminator and the updated generator into $\mathcal{D}$ and $\mathcal{G}$, respectively.}
		\ENDWHILE 
	\end{algorithmic}
\end{algorithm}

The following theorem provides the theoretical convergence guarantee for Fictitious GAN. It shows that assuming best response at each update in Fictitious GAN, the distribution of the mixture outputs from the generators converge to the data distribution. The intuition of the proof is that fictitious play achieves a Nash equilibrium in two-player zero-sum games. Since the optimal solution of GAN is a unique equilibrium in the game, fictitious GAN achieves the optimal solution. 
\begin{theorem}\label{convergence}
	Suppose the discriminator and the generator are updated according to the best-response strategy at each iteration in Fictitious GAN, then 
	\begin{align}
	\label{eq:pg_converge} &  \lim_{n \to \infty} \frac{1}{n} \sum_{w = 0}^{n-1} p_{g,w} (\bx) = p_d (\bx), \\
	\label{eq:d_converge} & \lim_{n \to \infty} D_{n} (\bx) = \frac{1}{2},
	\end{align}
	where $D_{w}(\bx)$ is the output from the $w$-th trained discriminator model and $p_{g,w}$ is the generated distribution due to the $w$-th trained generator. 
\end{theorem}

\subsection{Fictitious GAN as a Meta-Algorithm}

One advantage of Fictitious GAN is that it can be applied on top of existing GANs. 
Consider the following minimax problem:
\begin{align}\label{eq:ganvariant}
\min_{G} \max_{D} V(G,D) = \mathsf{E}_{\bx \sim p_d(\bx)} \{f_0 (D(\bx)) \} + \mathsf{E}_{\bz \sim p_z(\bz)} \{f_1(D(G(\bz)) ) \},
\end{align}
where $f_0(\cdot)$ and $f_1(\cdot)$ are some quasi-concave functions depending on the GAN variants. Table \ref{tab:generalize_gan} shows the family of \textit{f}-GAN \cite{chen2018training,nowozin2016f} and Wasserstein GAN.

We can model these GAN variants as two-player zero-sum games and the training algorithms for these variants of GAN follow by simply changing $f_0(\cdot)$ and $f_1(\cdot)$ in the updating rule accordingly in Algorithm \ref{alg1}. Following the proof in Theorem \ref{convergence}, we can show that the time average of generated distributions will converge to the data distribution and the discriminator will converge to $D^*$ as shown in Table \ref{tab:generalize_gan}.

\begin{table}
	\centering
	\caption{Variants of GANs under the zero-sum game framework.}
	\begin{adjustbox}{max width=0.8\textwidth,center}
	\begin{tabular}{|c|c|c|c|c|c|}
		\hline
		Divergence metric & $f_0 (D)$ & $f_1(D)$  & $D^{\ast}$ & value of the game  \\ \hline 
		Kullback-Leibler & $\log (D)$ & $1 - D$  & 1 & 0\\ \hline
		Reverse KL & $- D $ & $\log D $ & 1 & -1 \\ \hline
		Pearson $\chi^2$ & $D$ & $-\frac{1}{4} D^2 - D$ & 0 & 0 \\ \hline
		Squared Hellinger $\chi^2$ & $1-D$ & $1- 1/D$ & 1 & 0 \\ \hline
		Jensen-Shannon  & $\log (D) $ & $\log (1-D) $ &  $\frac{1 }{ 2}$ &-$\log 4$ \\ \hline
		WGAN & $D  $  & $-D$ & 0 & 0\\ 
		\hline
	\end{tabular}
	\end{adjustbox}
	\label{tab:generalize_gan}
\end{table}

\section{Experiments}

Our Fictitious GAN is a meta-algorithm that can be applied on top of existing GANs.
To demonstrate the merit of using Fictitious GAN, we apply our meta-algorithm on DCGAN~\cite{radford2015unsupervised} and its extension conditional DCGAN.
Conditional DCGAN allows DCGAN to use external label information to generate images of some particular classes.
We evaluate the performance on a synthetic dataset and three widely adopted real-world image datasets.
Our experiment results show that 
Fictitious GAN could improve visual quality of both DCGAN and conditional GAN models.

\textbf{Image dataset.}
(1) \textbf{MNIST}: contains 60,000 labeled images of 28 $\times $ 28 grayscale digits. 
(2) \textbf{CIFAR-10}: consists of colored natural scene images sized at 32 $\times$ 32 pixels. There are 50,000 training images and 10,000 test images in 10 classes.
(3) \textbf{CelebA}: is a large-scale face attributes dataset with more than 200K celebrity images, each with 40 attribute annotations.

\textbf{Parameter Settings.}
We used Tensorflow for our implementation.
Due to GPU memory limitation, we limit number of historical models to 5 in real-world image dataset experiments.
More architecture details are included in supplementary material.

\subsection{2D Mixture of Gaussian}
\begin{figure}[ht]
	\centering
	\begin{subfigure}[b]{0.19\textwidth}
		\centering
		\includegraphics[width=0.95\textwidth]{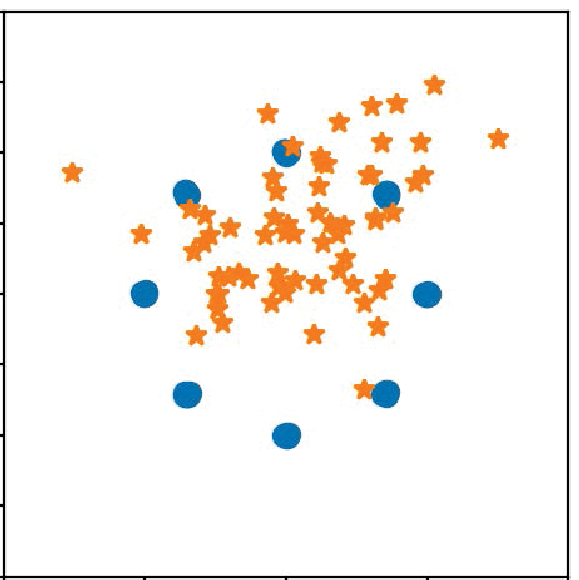}
		\caption*{Iteration 0}
		\label{fig:gauss2d_0}
	\end{subfigure}
	\begin{subfigure}[b]{0.19\textwidth}
		\centering
		\includegraphics[width=0.95\textwidth]{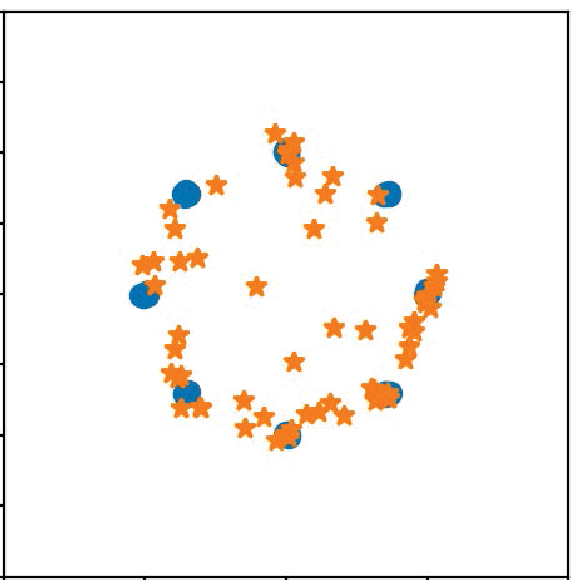}
		\caption*{Iteration 10k}
		\label{fig:gauss2d_10k}
	\end{subfigure}
	\begin{subfigure}[b]{0.19\textwidth}
		\centering
		\includegraphics[width=0.95\textwidth]{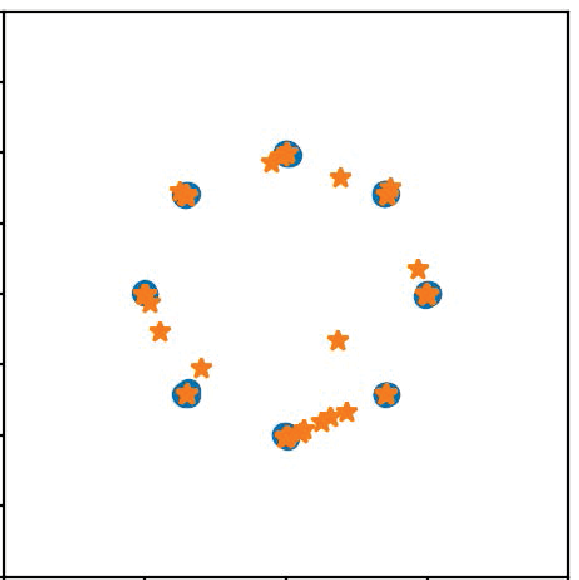}
		\caption*{Iteration 20k}
		\label{fig:gauss2d_20k}
	\end{subfigure}
	\begin{subfigure}[b]{0.19\textwidth}
		\centering
		\includegraphics[width=0.95\textwidth]{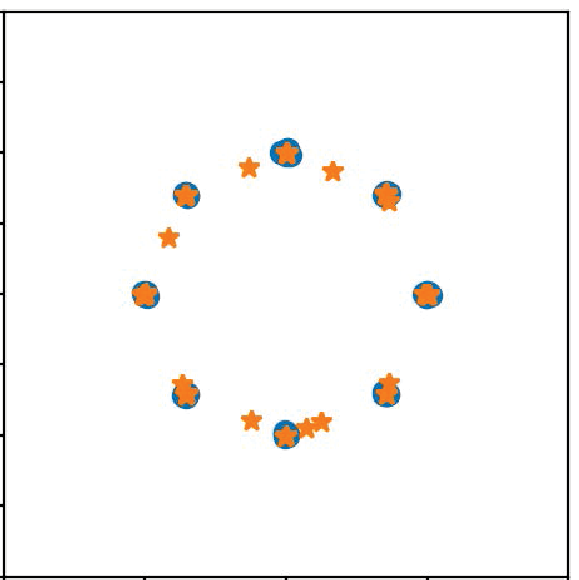}
		\caption*{Iteration 30k}
		\label{fig:gauss2d_30k}
	\end{subfigure}
	\begin{subfigure}[b]{0.19\textwidth}
		\centering
		\includegraphics[width=0.95\textwidth]{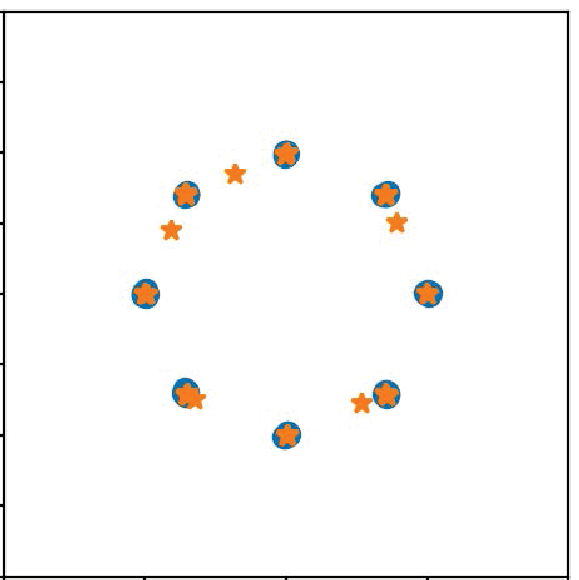}
		\caption*{Iteration 34k}
		\label{fig:gauss2d_34k}
	\end{subfigure}
	\caption{Performance of Fictitious GAN on 2D mixture of Gaussian data. The data samples are marked in blue and the generated samples are marked in orange.}
	\label{fig:gauss_example}
\end{figure}

%
%
Fig.~\ref{fig:gauss_example} shows the performance of Fictitious GAN for a mixture of 8 Gaussain data on a circle in 2 dimensional space. We use the network structure in~\cite{metz2016unrolled} to evaluate the performance of our proposed method. The data is sampled from a mixture of 8 Gaussians uniformly located on a circle of radius 1.0. Each has standard deviation of 0.02. The input noise samples are a vector of 256 independent and identically distributed (i.i.d.) Gaussian variables with mean zero and unit standard deviation.



While the original GANs experience mode collapse~\cite{nguyen2017dual,metz2016unrolled}, Fictitious GAN is able to generate samples over all 8 modes, even with a single discriminator asymptotically. 

\subsection{Qualitative Results for Image Generation}
\begin{figure}[ht]
	\centering
	\begin{subfigure}[b]{0.3\textwidth}
		\centering
		\includegraphics[width=1.0\textwidth]{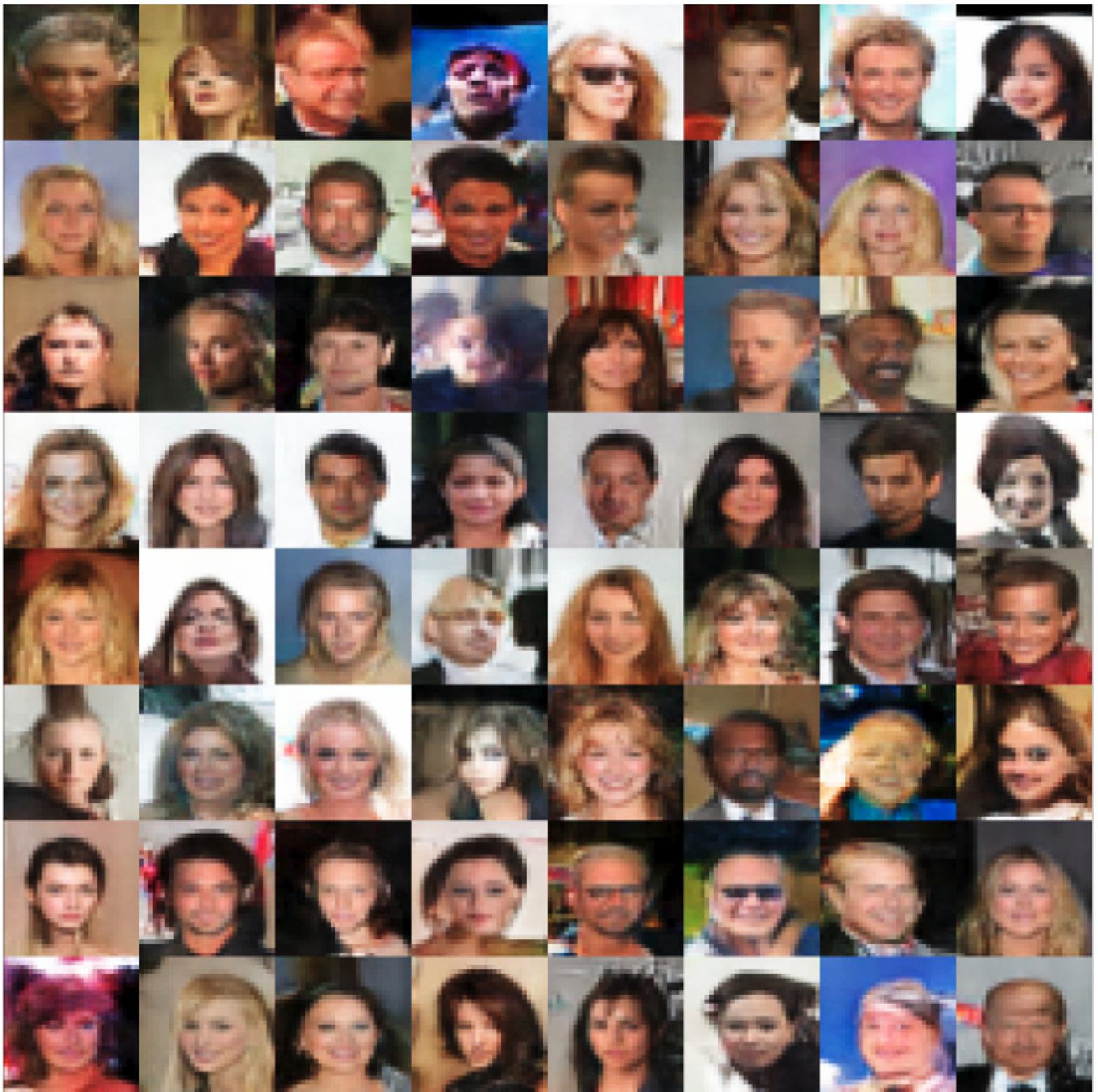}
	\end{subfigure}
	\begin{subfigure}[b]{0.3\textwidth}
		\centering
		\includegraphics[width=1.0\textwidth]{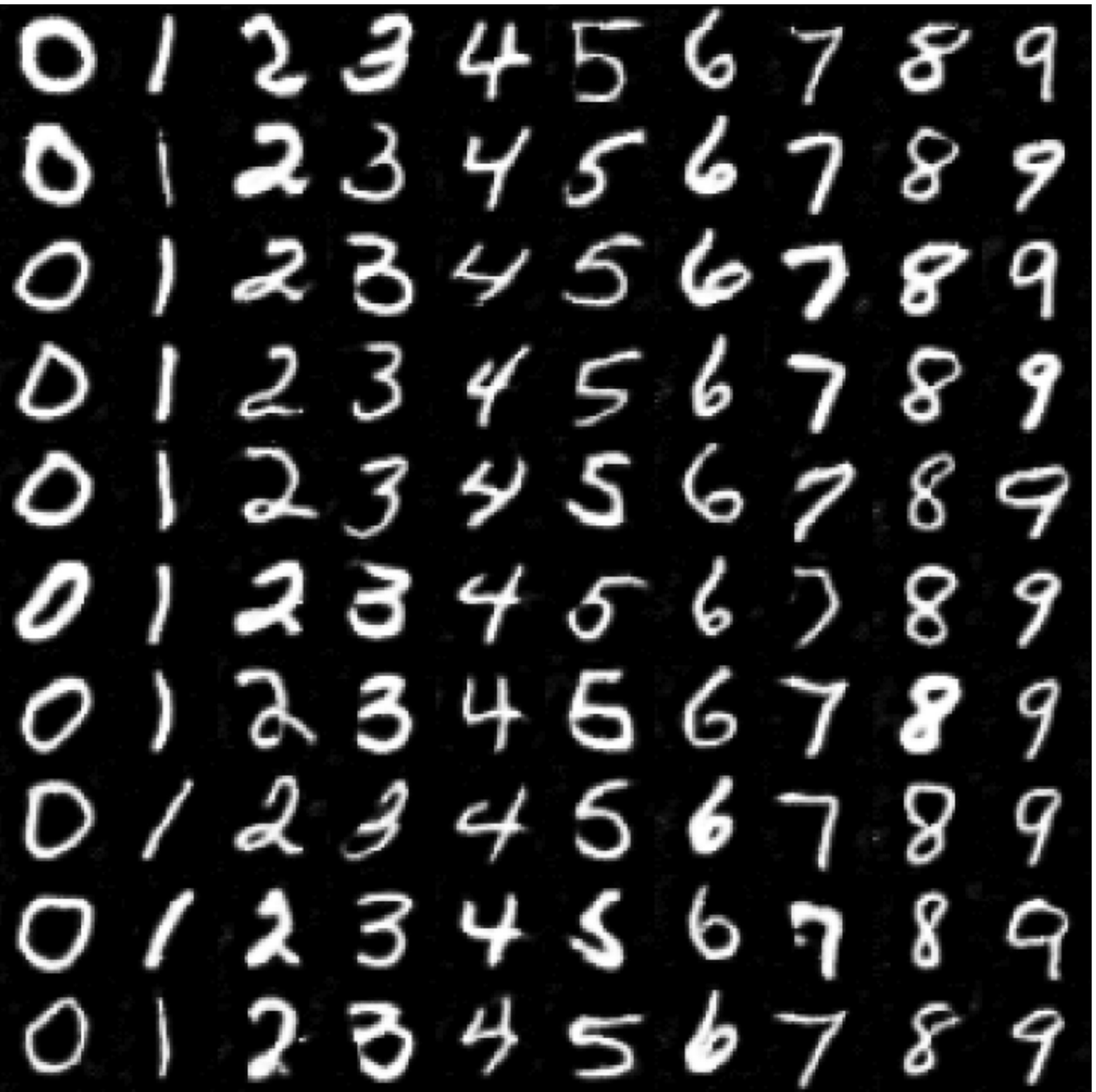}
	\end{subfigure}
	\begin{subfigure}[b]{0.3\textwidth}
		\centering
		\includegraphics[width=1.0\textwidth]{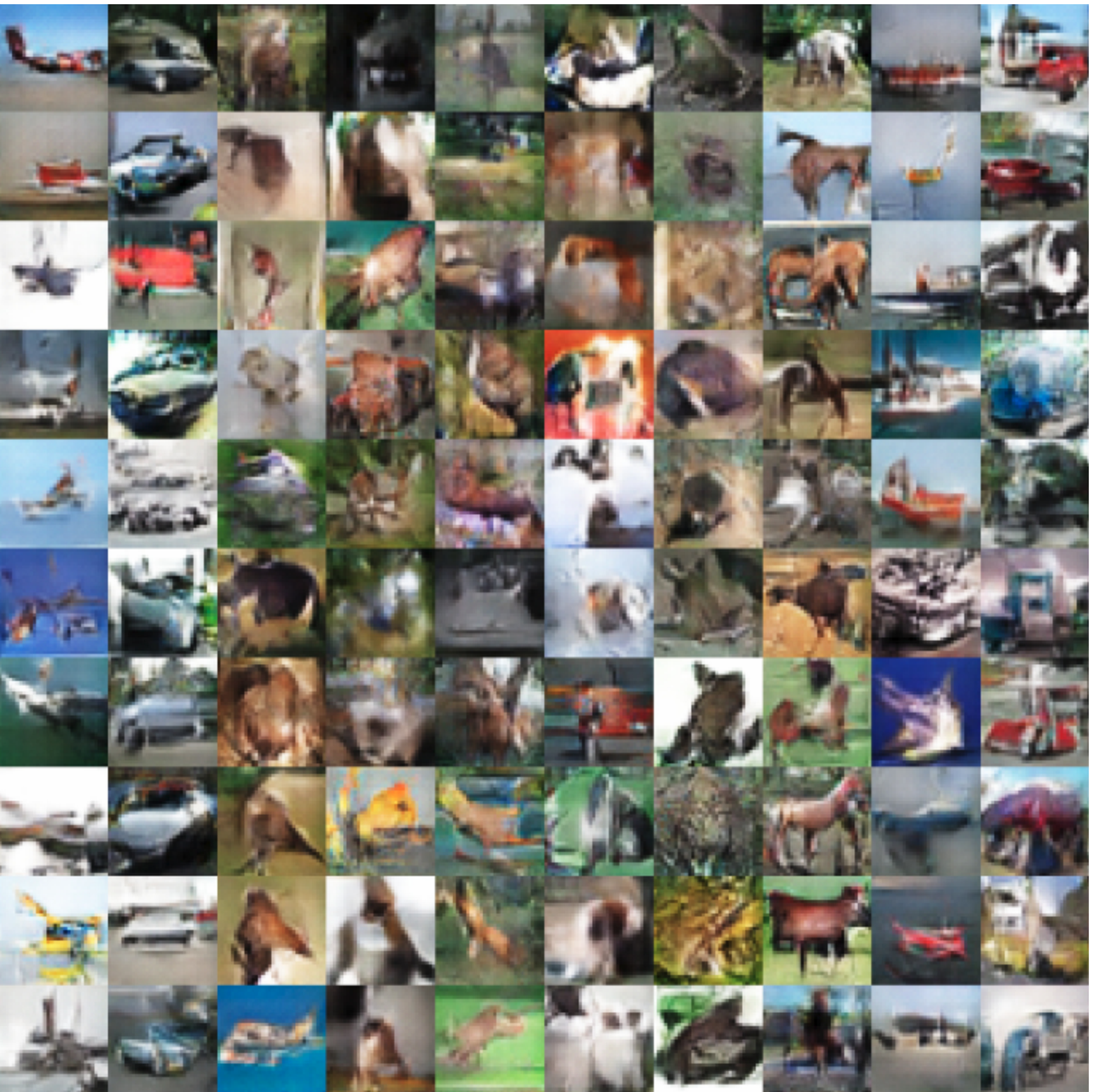}
	\end{subfigure}
	\\
	\begin{subfigure}[b]{0.3\textwidth}
		\centering
		\includegraphics[width=1.0\textwidth]{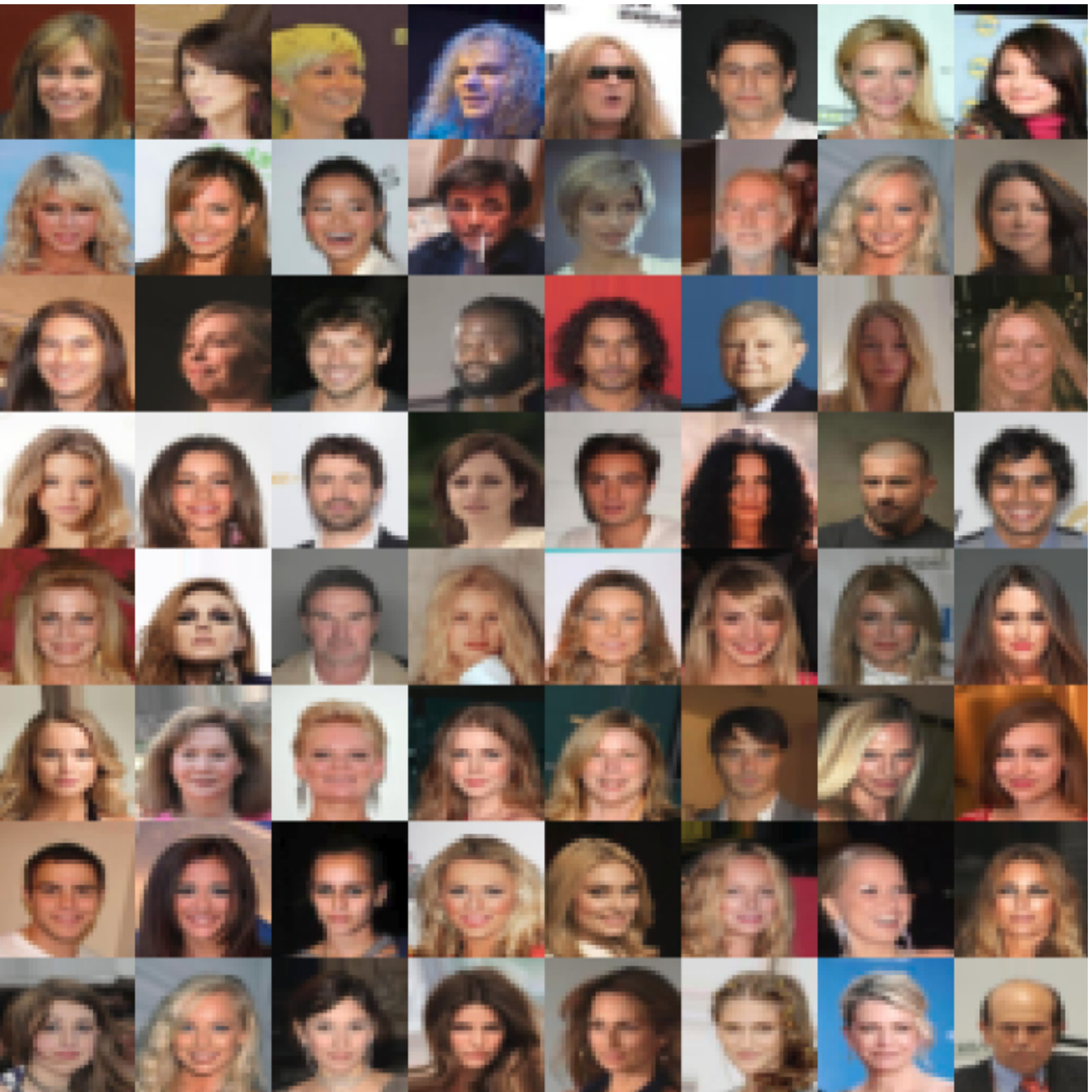}
	\end{subfigure}
	\begin{subfigure}[b]{0.3\textwidth}
		\centering
		\includegraphics[width=1.0\textwidth]{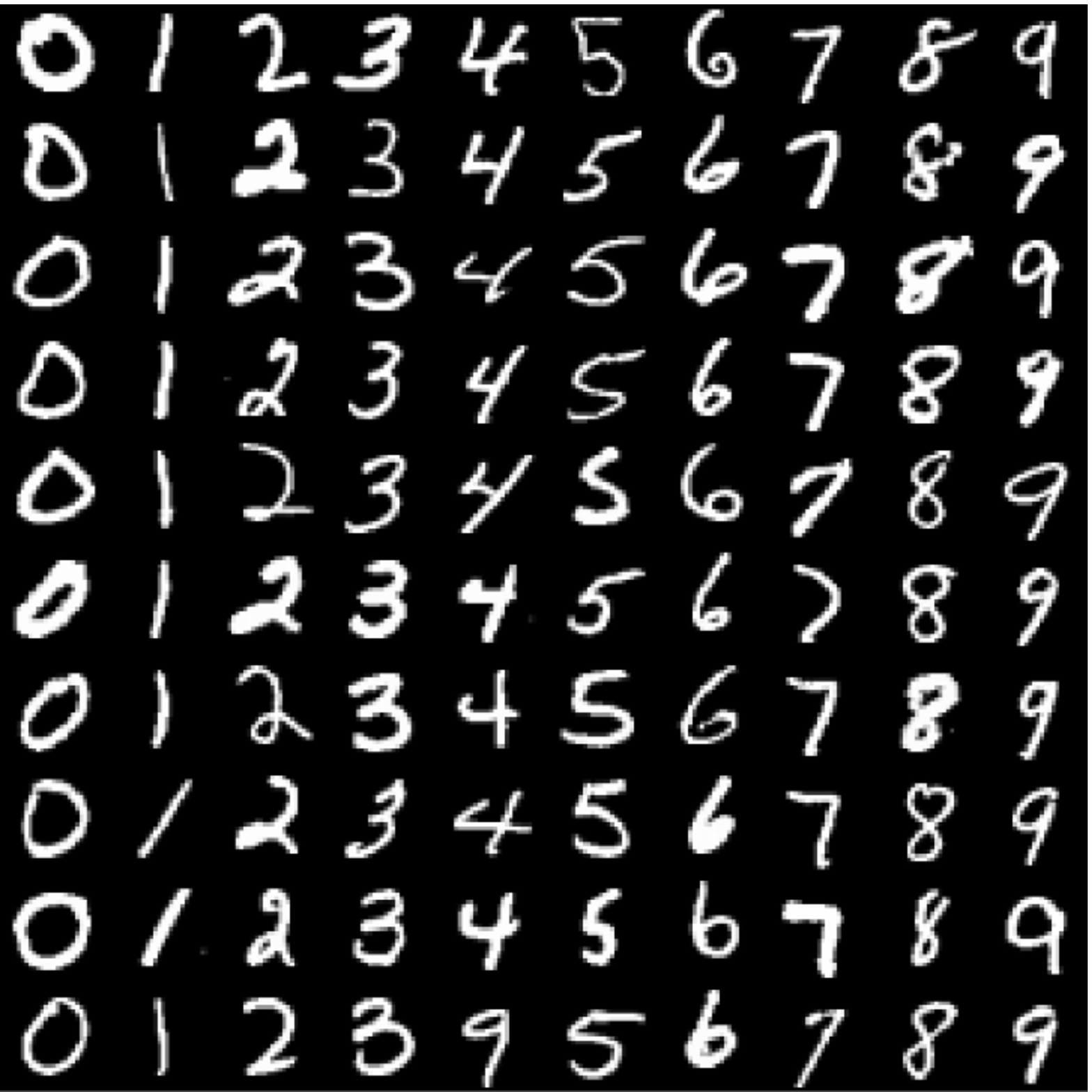}
	\end{subfigure}
	\begin{subfigure}[b]{0.3\textwidth}
		\centering
		\includegraphics[width=1.0\textwidth]{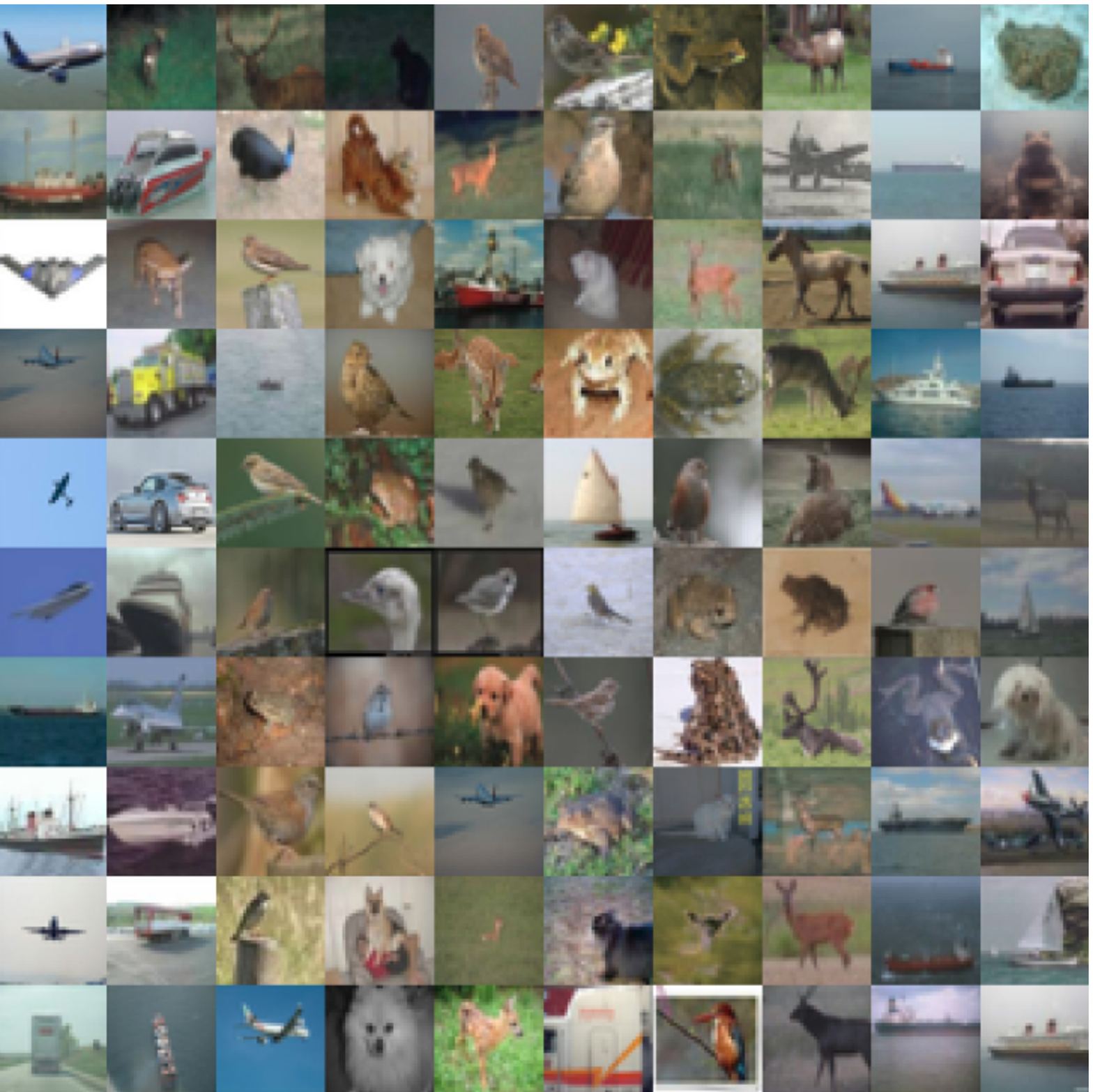}
	\end{subfigure}
	
	\caption{Generated images in CelebA, MNIST and CIFAR-10. Top row samples are generated, bottom row images are corresponding nearest neighbors in training dataset.}
	\label{Fig:generate_samples}
\end{figure}

We show visual quality of samples generated by DCGAN and conditional DCGAN, trained by proposed Fictitious GAN.
In Fig. \ref{Fig:generate_samples} first row corresponds to generated samples. We apply train DCGAN on CelebA dataset, and train conditional DCGAN on MNIST and CIFAR-10. Each image in the first row corresponds to the image in the same grid position in second row of Fig. \ref{Fig:generate_samples} . The second row shows the nearest neighbor in training dataset computed by Euclidean distance.
The samples are randomly drawn without cherry picking, they are representative of model output distribution.

In CelebA, we can generate face images with various genders, skin colors and hairstyles. 
In MNIST dataset, all generated digits have almost visually identical samples. Also, digit images have diverse visual shapes and fonts.
CIFAR-10 dataset is more challenging, images of each object have large visual appearance variance.  
We observe some visual and label consistency in generated images and the nearest neigbhors, especially in the categories of airplane, horse and ship.
Note that though we theoratical proved that Fictitious GAN could improve robustness of training in best response strategy, 
the visual quality still depends on the baseline GAN architecture and loss design, which in our case is conditional DCGAN.

%
\subsection{Quantitative Results}
In this section, we quantitatively show that DCGAN models trained by our Fictitious GAN could gain improvement over traditional training methods. Also, we may have a better performance by applying Fictitious gan on other existing gan models. The results of comparison methods are directly copied as reported.

\textbf{Metric.} The visual quality of generated images is measured by the widely used Inception score metric~\cite{salimans2016improved}. It measures visual objectiveness of generated image and correlates well with human scoring of the realism of generated images. Following evaluation scheme of ~\cite{salimans2016improved} setup, we generate 50,000 images from our model to compute the score.


\begin{table}[ht]
	\caption{Inception Score on CIFAR-10. }
	\begin{adjustbox}{max width=0.4\textwidth,center}
		\begin{tabular}{|c c|} \hline
			Method  & Score\\ \hline
			Fictitious cDCGAN* & \textbf{7.27 $\pm$ 0.10} \\ \hline
			DCGAN* \cite{huang2017stacked}(best variant) & 7.16 $\pm$ 0.10 \\ \hline
			MIX+WGAN* \cite{arora2017generalization} & 4.04 $\pm$ 0.07 \\ \hline 
			\hline
			Fictitious DCGAN   & 6.63 $\pm$ 0.06 \\ \hline
			DCGAN \cite{huang2017stacked} & \textbf{6.16 $\pm$ 0.07} \\ \hline
			GMAN \cite{durugkar2016generative} & 6.00 $\pm$ 0.19 \\ \hline
			WGAN \cite{arora2017generalization} & 3.82 $\pm$ 0.06 \\ \hline
			\hline
			Real data & 11.24 $\pm$ 0.12 \\ \hline
		\end{tabular}
	\end{adjustbox}
	\label{Tab::incept_score}
	\hspace{3cm} {Note: * denotes models that use labels for training.}
\end{table}

As shown in Table \ref{Tab::incept_score},
Our method outperforms recent state-of-the-art methods.
Specifically, we improve baseline DCGAN from $6.16$ to $6.63$; and conditional DCGAN model from $7.16$ to $7.27$. It sheds light on the advantage of training with the proposed learning algorithm.
Note that in order to highlight the performance improvement gained from fictitious GAN, the inception score of reproduced DCGAN model 
is 6.72, obtained without using tricks as \cite{salimans2016improved}.
Also, we did not use any regularization terms such as conditional loss and entropy loss to train DCGAN, as in \cite{huang2017stacked}. We expect higher inception score when more training tricks are used in addition to Fictitious GAN. 

%
\subsection{Ablation studies}
\begin{figure}[t]
	\centering
	\includegraphics[width=0.38\textwidth]{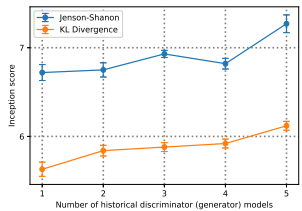}
	\caption{
	We show that Fictitious-GAN can improve Inception score as a meta-algorithm with larger number of historical models , 
	We select 2 divergence metrics from Table \ref{tab:generalize_gan}: Jenson-Shanon and KL divergence.}
	\label{fig:inception_trend}
\end{figure}

One hyperparameter that affects the performance of Fictitious GAN is the number of historical generator (discriminator) models. 
We evaluate the performance of Fictitious GAN with different number of historical models, and report the inception scores on the 150-th epoch in CIFAR-10 dataset in Fig. \ref{fig:inception_trend}.
We keep the number of historical discriminators the same as the number of historical generators.
We observe a trend of performance boost with an increasing number of historical models in 2 baseline GAN models.
The mean of inception score slightly drops for Jenson-Shannon divergence metric when the copy number is 4,
due to random initialization and random noise generation in training.

\section{Conclusion}

In this paper, we relate the minimax game of GAN to the two-player zero-sum game. This relation enables us to leverage the mechanism of fictitious play to design a novel training algorithm, referred to as fictitious GAN. In the training algorithm, the discriminator (resp. generator) is alternately updated as best response to the mixed output of the stale generator models (resp. discriminator). This novel training algorithm can resolve the oscillation behavior due to the pure best response strategy and the inconvergence issue of gradient based training in some cases. Real world image datasets show that applying fictitious GAN on top of the existing DCGAN models yields a performance gain of up to 8\%.

\section{Appendix}

\subsection{Proof of \eqref{eq:dynamic_example2}} \label{apendix8.1}
The eigenvalues of the transition matrix in \eqref{eq:trans} are $1+ \triangle i$ and $1-\triangle i$, and the corresponding eigenvectors are $\be_1 = [i,1]'$ and $\be_2 = [i,-1]'$, respectively. The initial condition $(x_0,y_0)$ can written as:
\begin{equation}\label{eq:initial}
\begin{bmatrix}
x_{0}    \\
y_{0}  
\end{bmatrix}
=
\frac{y_0-x_0i}{2}
\begin{bmatrix}
i    \\
1
\end{bmatrix}
+
\frac{-y_0-x_0i}{2}
\begin{bmatrix}
i    \\
-1
\end{bmatrix}
= \frac{y_0-x_0i}{2} \be_1 + \frac{-y_0-x_0i}{2} \be_2.
\end{equation}
Combining \eqref{eq:trans} and \eqref{eq:initial}, we have
\begin{equation}
\begin{bmatrix}
x_{n}    \\
y_{n}  
\end{bmatrix}
= (1+\triangle i)^n\frac{y_0-x_0i}{2} \be_1 + (1-\triangle i)^n\frac{-y_0-x_0i}{2} \be_2.
\end{equation}
Define $c_1 = \sqrt{(1+\triangle ^2)}$, $c_2 = \frac{1}{2}\sqrt{x_0^2+y_0^2}$, $\theta =\arctan(\triangle)$ and $\beta = -\arctan \frac{x_0}{y_0}$. Then $(x_n, y_n)$ can be calculated as:
\begin{equation}
\begin{bmatrix}
x_{n}    \\
y_{n}  
\end{bmatrix}
=
\begin{bmatrix}
-c_1^nc_2\sin(n\theta+\beta)    \\
c_1^nc_2\cos(n\theta+\beta)   
\end{bmatrix}.
\end{equation}

\subsection{Proof of Corollary~\ref{cor:ne_minimax}}

Suppose $(\mu_1^*,\mu_2^*)$ is a Nash equilibrium, then from the definition, we know 
\begin{align}
u({\mu_1,\mu_2^*}) \leq u(\mu_1^*,\mu_2^*) \leq u(\mu_1^*,\mu_2)
\end{align} 
for any $\mu_1 \in \Delta S_1$ and $\mu_2\in \Delta S_2$. Hence we have
\begin{align}
\bar{v} &= \inf_{\mu_2 \in \Delta S_2}\sup_{\mu_1\in\Delta S_1}u(\mu_1,\mu_2) \\
&\leq \sup_{\mu_1\in\Delta S_1}u(\mu_1,\mu_2^*) \\
& \leq u(\mu_1^*,\mu_2^*) \\
&= \inf_{\mu_2\in \Delta S_2} u(\mu_1^*,\mu_2) \\
& \leq \sup_{\mu_1\in \Delta S_1}\inf_{\mu_2 \in \Delta S_2}u(\mu_1,\mu_2) \\
&=\underline{v}.
\end{align}
Since $v = \bar{v} =\underline{v}$, we obtain  $u(\mu_1^*,\mu_2^*) =v$.    

\subsection{Proof of Nash Equilibrium for Example 2}

Now, we show the mixed strategy $(\sigma_1^*,\sigma_2^*)$ that both players choose 10 and -10 with probability $\frac{1}{2}$ is a Nash equilibrium for the minimax game.

Take player 1 for instance, given $\sigma_2^*$, then for any possible mixed strategy $\sigma_1$, where $\sigma_1(x)$ indicates the probability she chooses value x, we know the expected utility for her is:
\begin{equation}
\frac{1}{2}\int_{x=-10}^{10}10*x\sigma_1(x) dx + \frac{1}{2}*\int_{x=-10}^{10}(-10)*x\sigma_1(x) dx = 0.
\end{equation}
Hence no matter what her strategy is, the expected utility is always 0 and therefore player 1 has no incentive to deviate from strategy $\sigma_1^*$ given $\sigma_2^*$. Similarly, we can show player 2 has no incentive to deviate from $\sigma_2^*$ given $\sigma_1^*$. Thus, the mixed strategy $(\sigma_1^*,\sigma_2^*)$ is a Nash equilibrium.

\subsection{Proof of Theorem~\ref{theorem:gan_ne}}

Let $V(p_g,D)$ be as defined in \eqref{eq:gan_pg}.
A Nash equilibrium in a zero-sum game is a mixed strategy $(\mu_D^*, \mu_G^*)$ with corresponding pdf of $(\sigma_D^*, \sigma_g^*)$ such that
\begin{align}
\label{eq:d_mix} \sigma_D^* = \arg\max_{\sigma_D} \int_D \int_g \sigma_D \sigma_g^* V(p_g, D) d g dD\\
\label{eq:g_mix} \sigma_g^* = \arg\min_{\sigma_g} \int_D \int_g \sigma_g \sigma_D^* V(p_g, D) dg dD.
\end{align}

Define $p^*_g (\bx) = \int_g \sigma^*_g p_g (\bx) dg$. Note that $p^*_g (\bx) $ is also a valid probability density over $\bx$. We have 
\begin{align}
\int_g \sigma_g^* V( p_g, D) dg &= \int_g \sigma_g^* V(p_g, D) dg\\
& =  \int_g \sigma_g^* \left[  \mathsf{E}_{\bx \sim p_d(\bx)} \{\log D(\bx) \} + \mathsf{E}_{\bx \sim p_g(\bx)} \{\log (1- D(\bx) ) \} \right] dg \\
&  =  \mathsf{E}_{\bx \sim p_d(\bx)} \{\log D(\bx) \} + \mathsf{E}_{\bx \sim p^*_g(\bx)} \{\log (1- D(\bx) ) \\
& = V(p^*_g, D).
\end{align}
Hence \eqref{eq:d_mix} can be rewritten as:
\begin{equation}
\sigma_D^* = \arg\max_{\sigma_D} \int_D  \sigma_D  V(p_g^*, D)  dD,
\end{equation}
and given $p^*_g$, the optimal strategy for the discriminator is to choose $D^* (\bx) = \frac{p_d(\bx)}{p_d (\bx)+p^*_g (\bx)}$ with probability 1, which means the best response is a pure strategy.

Therefore, at any Nash equilibrium, the generator generates data following a pure distribution $p_g^*$, while the discriminator chooses a pure response $D^* (\bx)$. Moreover, $p_g^* = p_d$ is the only solution to \eqref{eq:g_mix}, i.e., the generator has no incentive to deviate. Consequently, the only possible Nash equilibrium is $p_g^* = p_d$ and $D^* (\bx) =\frac{1}{2} $ for any $\bx$.

\subsection{Proof of Theorem~\ref{convergence}}

\begin{proof}
	For the minimax game of~\eqref{eq:gan}, let $p_g(\bx)$ be the generated distribution. We rewrite the optimization problem as
	\begin{align}\label{eq:gan_pg}
	\min_{p_g} \max_{D} V(p_g,D) = \mathsf{E}_{\bx \sim p_d(\bx)} \{\log D(\bx) \} + \mathsf{E}_{\bx \sim p_g(\bx)} \{\log (1- D(\bx) ) \}.
	\end{align}
	With $p_g$ fixed, $V(p_g, D)$ is semi-continuous and quasi-concave in $D$; and with $D$ fixed, $V(p_g, D)$ is semi-continuous and quasi-convex in $p_g$. Thus, the utility function $V(p_g,D)$ satisfies the conditions in Theorem~\ref{theorem:sion}. 
	
	Moreover, by Theorem~\ref{theorem:gan_ne}, the unique Nash equilibrium of the game is shown to be $p_g^{\ast}(\bx) = p_d (\bx)$ and $D^{\ast} (\bx) = 1/2$ for all $\bx$. The value of the game is $V(p_g^{\ast}(\bx) , D^{\ast} (\bx) ) = - \log 4$.
	
	In Fictitious GAN, the discrimination function of $n$-th model satisfies $D_n = \arg\max_D \frac{1}{n} \sum_{w=0}^{n-1} V(p_{g,w}, D) $.
	Let $\bar{p}_{g,n} = \frac{1}{n}\sum_{w=0}^{n-1}  p_{g,w}$. It is easy to see that $\bar{p}_{g,n} $ is also a valid pdf. Then we have
	\begin{align}
	\frac{1}{n}  \sum_{w=0}^{n-1}  V(p_{g,w}, D)  &= \mathsf{E}_{\bx \sim p_d(\bx)} \{\log D(\bx) \} +\frac{1}{n}\sum_{w=0}^{n-1}  \mathsf{E}_{\bx \sim p_{g,w}(\bx)} \{\log (1- D(\bx) ) \} \\
	&= V(\bar{p}_{g,n} , D).
	\end{align}
	Therefore, the optimal discrimination function of $n$-th model is calculated as 
	\begin{align}\label{eq:Dn}
	D_n (\bx) = \frac{p_d (\bx)}{p_d (\bx) + \bar{p}_{g,n} (\bx)}.
	\end{align}
	Thus, $V(\bar{p}_{g,n} , D_n) = 2JSD(\bar{p}_{g,n} || p_d) - \log 4 $, where $JSD(p_g|| p_d)$ is the Jensen-Shannon divergence between $p_g$ and $p_d$ as defined in~\cite{goodfellow2014generative}. By Theorem~\ref{theorem:fict_play_game}, we have 
	$\frac{1}{n}  \sum_{w=0}^{n-1}  V(p_{g,w}, D_n) \to - \log 4$,
	which implies that $JSD(\bar{p}_{g,n} || p_d) $ tends to zero. Since $JSD(p_g|| p_d)  = 0$ if and only if $p_g(\cdot) = p_d (\cdot)$, \eqref{eq:pg_converge} is established. Combining \eqref{eq:Dn} and \eqref{eq:pg_converge} yields \eqref{eq:d_converge}.
	
\end{proof}

\subsection{Network Architectures and Parameters}
All architectures are chosen as recommended by a publicly avaiable implementation\footnote{https://github.com/carpedm20/DCGAN-tensorflow}.

\textbf{Experiment on synthetic data:}
The generator has two hidden layers of size 128 with ReLU activation. The last layer is a linear projection to two dimensions. The discriminator has one hidden layer of size 128 with ReLU activation followed by a fully connected network to a sigmoid activation. 
All the biases are initialized to be zeros and the weights are initalilzed via the ``Xavier" initialization~\cite{glorot2010understanding}. 
The training updates the discriminator and the generator using 3 sub-iterations. The Adam optimizer is used to train the discriminator with 2e-4 learning rate and the generator with $1.2\times 10^{-4}$ learning rate. The minibatch sample number is 64. 

\textbf{Experiment on MNIST, CIFAR-10, Celeb-A:}
GAN networks were trained using the Adam optimizer~\cite{kingma2014adam} with batches of size 64 and learning rate $2\times 10^{-4}$, for around 150K generator iterations in the case of CIFAR-10 and 100K for MNIST.

\bibliographystyle{splncs}
\bibliography{all_bib}
\end{document}